# Interpreting Structured Perturbations in Image Protection Methods for Diffusion Models

Michael R. Martin[1*], Garrick Chan[1], Dr. Kwan-Liu Ma[1]
[1]University of California, Davis

*Corresponding Author: csemartin@ucdavis.edu

**Abstract**
Recent image protection mechanisms such as Glaze and Nightshade introduce imperceptible, adversarially designed perturbations intended to disrupt downstream text-to-image generative models. While their empirical effectiveness has been demonstrated, the internal structure, detectability, and representational behavior of these perturbations remain poorly understood. In this study, we demonstrated a systematic explainable AI analysis of image protection perturbations using a unified framework that integrates white-box feature-space inspection and black-box signal-level probing. Through latent-space clustering, feature-channel activation analysis, occlusion-based spatial sensitivity mapping, and frequency-domain spectral characterization, we revealed that modern protection mechanisms operate as structured, low-entropy perturbations that remain tightly coupled to underlying image content across representational, spatial, and spectral domains in all evaluated cases. We showed that protected images preserve content-driven feature organization with protection-specific substructure rather than inducing global representational drift. Detectability is governed by interacting effects of perturbation entropy, spatial deployment, and frequency alignment as revealed through combined synthetic and spectral analyses, with sequential protection amplifying detectable structure rather than suppressing it. Frequency-domain analysis further demonstrated that Glaze and Nightshade redistribute energy along dominant image-aligned frequency axes rather than introducing spectrally diffuse noise. These results suggested that contemporary image protection operates through structured feature-level deformation rather than semantic dislocation, providing mechanistic insight into why protection signals remain visually subtle yet consistently detectable. This work advances the interpretability of adversarial image protection and informs the design of future defenses and detection strategies for generative AI systems.

# 1. Introduction
Text-to-image diffusion models have rapidly transformed visual content creation by enabling users to synthesize high-quality imagery directly from natural language prompts. Systems such as Stable Diffusion, DALL·E, and Midjourney now support a diverse set of generated artistic styles such as photorealistic rendering, stylized illustration, and conceptual design at unprecedented scale. These models are increasingly embedded into commercial, artistic, and scientific workflows, reshaping how visual media is produced, distributed, and valued. Despite their significant success, diffusion models rely on massive, web-scale training datasets compiled largely through automated scraping (Rombach et al., 2022). These datasets contain image–text pairs ranging from hundreds of millions to billions, many of which originate from artists who had not consented to their work being used for machine learning. This practice had generated widespread concern regarding intellectual property rights, economic displacement, environmental cost, and the erosion of creative labor protections. Artists routinely publish their work online for professional exposure, community engagement, and income generation. However, these same online platforms have become primary data sources for large-scale AI training pipelines. In parallel, model personalization and fine-tuning techniques demonstrate that only a small number of images are sufficient for a diffusion model to learn and reproduce a specific artistic style. Public reports of AI systems generating images that replicate composition, signature patterns, and even visible watermarks further intensify concerns over unauthorized appropriation of creative identity.

In response to these risks, a class of image protection tools has emerged that intentionally modifies images prior to online publication. These tools introduce carefully engineered perturbations that remain largely imperceptible to human viewers but are intended to disrupt downstream model training. This strategy represents a fundamental shift in how creative content is defended in the age of generative AI: images are no longer passive artistic artifacts, but active adversarial signals embedded into data pipelines. However, the protection ecosystem is inherently adversarial. As protection methods advance, countermeasures designed to detect, reconstruct, or neutralize these perturbations are developed in parallel. This creates a rapidly escalating arms race between protection and purification systems, unfolding entirely within the opaque representational space of deep generative models.

While prior studies demonstrate that both protection and purification methods can succeed empirically, far less is known about how these systems operate internally, what features they exploit, and why certain perturbations persist while others may not be as successful. Explainable Artificial Intelligence (XAI) offers a critical framework for addressing this gap. XAI seeks to make the internal representations, feature responses, and decision logic of deep neural networks accessible to human interpretation. While XAI has been widely applied in domains such as medical imaging, security, and fairness analysis, its systematic application to generative model protection methods and, subsequently, perturbation purification methods remains limited. Furthermore, understanding the internal mechanisms by which generative and detection models perceive protected images is critical for developing defenses that are not only empirically effective but also scientifically interpretable.

This study is motivated by the need to shift the evaluation of artist-protection methods from purely outcome-based success metrics toward an interpretable, signal-structure–oriented understanding of perturbation behavior. Rather than treating protection as a black-box trick that either works or fails, we view perturbations as structured signals whose effects can be analyzed across feature space, activation space, and frequency space. This paper introduces a unified, explainability-driven framework for analyzing modern artist-protection perturbations in diffusion-based generative pipelines. We combine internal model interpretability with external, model-agnostic signal analysis to reveal how protected images are represented, detected, and transformed across multiple computational layers. This perspective enables a deeper understanding of why certain defenses succeed, why purification techniques generalize across protection styles, and where the fundamental limits of current protection strategies lie. This work aims to establish a principled foundation for the next generation of artist-protection technologies, defenses that are transparent, analyzable, and resilient against future purification systems, by grounding generative model security in interpretable representations rather than heuristic obfuscation.

## 2. Background and Related Work

This section situates our study within prior work on perturbation-based artist protection, countermeasures targeting such protections, and explainable AI (XAI) techniques for understanding perturbations in image models. We focus specifically on mechanisms and empirical findings that inform our later problem formulation and analysis, rather than on ethical or socio-technical context.

### 2.1 Perturbation-Based Data Poisoning for Generative Models

Adversarial methods, such as Glaze and Nightshade (Figure 1), can be understood as data perturbation methods on text-to-image pipelines: they deliberately modify training images so that models trained on these data acquire systematically distorted internal representations (Liang et al., 2023; Shan et al., 2023a; Shan et al., 2024). Unlike classic adversarial examples, which typically optimize perturbations for a fixed classifier and a single decision, these techniques target future training of large diffusion models under unknown architectures and hyperparameters. Formally, let $x \in R^{H \times W \times 3}$ denote an artist's image, $y$ denote associated text (e.g., a caption or artist identifier), and $f_\theta$ a generative model trained on a dataset $D = \{(x_i, y_i)\}$. Protective perturbation methods construct a modified image $\tilde{x} = x + \delta$ subject to a perceptual constraint $\|\delta\|_{perceptual} \leq \epsilon$, with the goal that models trained on $(\tilde{x}, y)$ learn an altered mapping from textual concepts or style tokens to visual features. The design problem is therefore to identify perturbations that are simultaneously: **(1)** Low-distortion in perceptual metrics (so that artists can share them), **(2)** High-impact on downstream learned representations, and **(3)** Robust against transformations, filtering, and targeted purification. This "train-time adversarial objective" distinguishes artist-protection perturbations from most adversarial robustness literature, which assumes inference-time access to a deployed model and does not typically consider long-term data reuse in large-scale generative training.

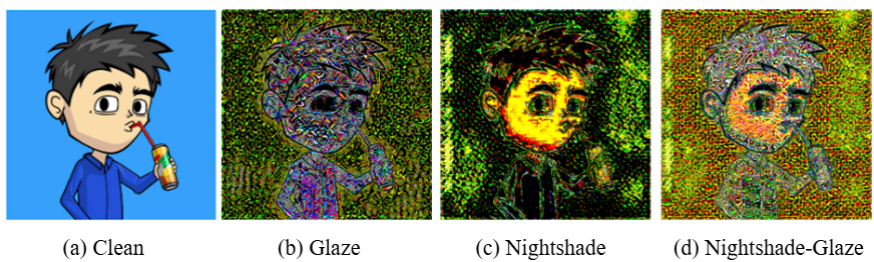

(a) Clean  (b) Glaze  (c) Nightshade  (d) Nightshade-Glaze

*Figure 1: Comparison of perturbation patterns produced by Glaze and Nightshade on the same clean input image. Panel (a) shows the original image. Panels (b–d) visualize the corresponding perturbations obtained by computing the pixelwise difference between the clean image and each protected output, with the difference contrast-enhanced for visibility.*

## 2.2 Protection Tools
### 2.2.1 Glaze: Style-Space Cloaking
Glaze is designed to prevent text-to-image systems from faithfully imitating an artist's style by operating primarily in a deep feature space rather than raw pixels (Shan et al., 2023a). The method first analyzes the artist's style of a piece $X$ in feature space and compares it to that of the styles of artists whose work has entered the public domain. A target style $T$ is chosen such that it is neither too similar to the user's style (which would offer weak protection) nor too different (which would create too much visual noise). Next, using a pre-trained feature extractor and style transfer model, $X_T$ is generated under the hood, with the colors and composition mostly intact. Then, $X$ is perturbed towards $X_T$ such that the distance between $X$ and $X_T$ in feature space is maximized while the visual difference remains below a user-specified threshold. Empirically, Glaze has been shown to substantially distort style reconstructions in downstream diffusion models, even when only a small number of Glazed artworks are used during training, and to retain some effectiveness under compression and moderate rescaling (Shan et al., 2023b). However, subsequent analyses highlight that Glaze's perturbations exhibit structured, low-entropy patterns that recur across images, making them amenable to reconstruction and removal by purification models that explicitly target such regularities (Foerster et al., 2025).

### 2.2.2 Mist: Denoising-Process Misdirection
Mist takes a different approach by directly targeting the iterative denoising process of diffusion models (Liang et al., 2023). Unlike Glaze, which primarily manipulates style embeddings, Mist constructs perturbations such that, when the protected image is used during training, the model's learned reverse-diffusion dynamics favor unlikely or unstable trajectories in latent space. At a high level, Mist seeks perturbations $\delta$ that maximize the discrepancy between the model's ideal denoising path and the path induced by $\tilde{x} = x + \delta$, while respecting a perceptual constraint on $\delta$. In practice, this is implemented by differentiating through the diffusion steps and penalizing high-probability, visually faithful reconstructions during optimization (Liang et al., 2023). This denoising-centric design improves robustness against some forms of post-processing, but at the cost of more visually noticeable artifacts than Glaze, especially at higher protection strengths. Thus, Mist presents a trade-off where it aims for stronger disruption of generative behavior, but is less tuned to artists' tolerance for visible perturbations and has not undergone the same degree of user-centered evaluation.

### 2.2.3 Nightshade: Concept-Space Corruption

Nightshade generalizes from style protection to concept poisoning by misaligning specific textual concepts with their visual counterparts (Shan et al., 2024). Given a human- and classifier-aligned concept $C$ (e.g., "dog"), Nightshade identifies an unrelated concept $A$ (e.g., "cat"), and constructs perturbations such that protected images remain recognizable as $C$ to humans and standard classifiers but are drawn toward the representation of $A$ in feature space used during training. Operationally, Nightshade uses a generative backbone (e.g., Stable Diffusion) to synthesize an auxiliary sample for concept $A$, then perturbs the original image towards this auxiliary target in a latent or feature space with a constrained pixel-space distortion budget (Shan et al., 2024). Because many concepts in datasets such as LAION-Aesthetic are supported by relatively few images, poisoning even tens of samples can significantly bias the learned mapping from the text token for $C$ to its visual representation (Schuhmann et al., 2022; Shan et al., 2024). An important observation from Nightshade's evaluation is that concept poisoning often exhibits *spillover*: poisoning a single base concept (e.g., "dog") can induce collateral degradation for semantically related concepts (e.g., "wolf", "puppy") due to shared structure in embedding space (Shan et al., 2024). This suggests that perturbation patterns interact nontrivially with the geometry of semantic representations learned by diffusion models, a phenomenon that our later XAI analysis seeks to make explicit.

## 2.3 Counter-Protection and Purification
### 2.3.1 Diffusion-Based Reconstruction: IMPRESS

IMPRESS interrogates the question of how robust imperceptible perturbations are when an adversary has access to a strong generative prior over natural images (Cao et al., 2023). Given a candidate protected image $x_{ptb}$, IMPRESS searches for a purified image $x_{pur}$ that both remains close to $x_{ptb}$ in pixel space and lies near the range of a diffusion autoencoder $D(E(x_{pur}))$. The second term encourages consistency with the generative model's feature space, effectively penalizing high-frequency, feature-distribution outliers. Experiments show that this process can partially suppress Glaze-like perturbations and recover stylistic fidelity in reconstructed images (Cao et al., 2023). However, the IMPRESS method's real-world impact is contested, as some follow-up work argues that IMPRESS is less effective against carefully tuned protection settings or when applied under realistic scraping assumptions (Shan et al., 2023c). Despite this, IMPRESS demonstrates how access to diffusion priors significantly alters the threat landscape for perturbation-based defenses.

### 2.3.2 Noisy Upscaling and Style-Dependent Failure Modes

This process involves systematically evaluating several purification strategies, with a focus on noisy upscaling pipelines: adding Gaussian noise to the (potentially protected) image and then applying a super-resolution model trained to invert such noise. Because many protection methods concentrate energy in high-frequency bands, this noise–upscale cycle can substantially weaken perturbations while preserving the subjective appearance of the artwork. A notable finding is that purification efficacy depends strongly on artistic style, texture density, and color structure. Highly textured or noisy artworks tend to mask perturbations more effectively, while flat-shaded or low-texture works yield cleaner separation between perturbation and content, making them easier to purify. In some regimes, human raters could not reliably distinguish purified outputs from the original unprotected artworks, highlighting an asymmetry: purification can exploit strong priors and extensive compute, whereas artists are constrained by visibility and usability constraints on perturbations.

### 2.3.3 LightShed: Entropy-Based Detection and Removal

LightShed introduces a dedicated detection and purification architecture built around the assumption that protection methods produce perturbations with lower entropy and more regular structure than the underlying images they modify (Foerster et al., 2025). It trains an autoencoder so that the encoder suppresses semantic content and the decoder reconstructs perturbation patterns. Lightshed's method aims to detect, reconstruct, and remove perturbations from protected images in this manner. For an input image $x$, the autoencoder produces a reconstructed perturbation $\hat{\delta}$; the Shannon entropy $H(\hat{\delta})$ is computed over the reconstructed perturbation map. If $H(\hat{\delta})$ exceeds a learned or hand-tuned threshold, the image is flagged as protected, and the reconstruction is subtracted from the original to yield a "cleaned" version: $x_{clean} = x - \hat{\delta}$. LightShed achieves high reconstruction fidelity across several protection families, including Glaze and Nightshade-like perturbations, even when protections are composed or combined with moderate post-processing (Foerster et al., 2025). From a signal perspective, this success underscores that existing protections tend to occupy a relatively low-dimensional, structured subspace of possible perturbations. This observation motivates our experimental exploration of higher-entropy, spatially non-uniform, and frequency-adaptive designs.

## 2.4 Explainable AI and Spectral Perspectives on Perturbations

XAI for vision has traditionally focused on classification models, providing tools to inspect internal features, spatial attributions, and concept-level representations. Gradient-based localization methods such as Grad-CAM highlight regions that contribute strongly to a specific prediction by combining spatial feature maps with class gradients (Selvaraju et al., 2017). Axiomatic attribution methods like Integrated Gradients accumulate gradients along a path from a baseline to the input, yielding feature importance scores that satisfy rigorously defined properties (Sundararajan et al., 2017). Concept-based explanation frameworks such as TCAV probe internal representations using directional derivatives along concept vectors, linking neurons or subspaces to human-interpretable categories rather than individual pixels (Ghorbani et al., 2019). Feature visualization studies, including deconvolution-based methods and activation maximization, have revealed that individual channels encode increasingly complex structures, from oriented edges to textures and object parts, in convolutional networks (Zeiler & Fergus, 2014; Olah et al., 2017). In parallel, several works adopt a frequency-domain viewpoint on robustness. Yin et al. (2019) demonstrate that convolutional networks exhibit characteristic sensitivity patterns across spatial frequencies and that adversarial perturbations often concentrate energy in specific high-frequency regions. Spectral analysis has been used to diagnose brittle feature reliance and to design frequency-filtering defenses or regularizers that encourage models to align with more robust, low-frequency structures (Yin et al., 2019).

However, the intersection of XAI, spectral analysis, and data poisoning for generative models remains underexplored. Most evaluations of artist-protection methods emphasize output quality and empirical resistance to specific countermeasures, while most XAI studies focus on classification rather than generative or purification architectures (Foerster et al., 2025). There is comparatively little work that treats protection perturbations as analyzable signals in both spatial and frequency domains, connects those signals to internal activations and latent-space structure of detection models, and uses XAI methodology to characterize how purification systems exploit low-entropy patterns to reconstruct or remove perturbations. The present study addresses this gap by combining latent clustering, layer-wise activation analysis, occlusion-style spatial probing, and Fourier-domain characterization to obtain a multi-view explanation of how perturbations produced by contemporary artist-protection schemes are represented and processed by a state-of-the-art detection and purification model.

## 3. Problem Formulation and Research Questions
### 3.1 Problem Formulation

Let $D = \{(x_i, y_i)\}_{i=1}^{N}$ denote a large-scale image–text training dataset used to train a text-to-image diffusion model, where each image $x_i \in R^{H \times W \times 3}$ is paired with a text descriptor $y_i$. Artist-protection methods modify a subset of images $\{x_j\} \subset D$ to produce perturbed versions $\tilde{x}_j = x_j + \delta_j$, where $\delta_j$ is constrained to remain visually imperceptible under human perception. Unlike inference-time adversarial examples, these perturbations are designed to influence future model training, not the predictions of a fixed deployed network. The optimization objective of protective perturbations is therefore indirect: rather than maximizing immediate output error, they aim to induce persistent distortion in the learned latent representations of a generative model trained on $D \cup \{(\tilde{x}_j, y_j)\}$. At the same time, purification and detection systems introduce a second optimization objective. Let $P(x)$ denote the output of a purification operator applied to an image "x" prior to training. The adversarial game becomes triadic:

**Protection Objective -** Construct $\tilde{x} = x + \delta$ such that x and $\tilde{x}$ are perceptually indistinguishable to humans, and models trained on $\tilde{x}$ learn misaligned style or concept representations.

**Purification Objective -** Given $\tilde{x}$, estimate $\hat{\delta}$ such that $x_{clean} = \tilde{x} - \hat{\delta}$ approximates the original unperturbed image $x$, while minimizing false positives.

**Detection Objective -** Decide whether an input image contains a perturbation based on measurable signal properties (e.g., entropy, frequency distribution, or learned reconstruction error).

All three processes operate within high-dimensional, non-linear representation spaces, and none of them observes the internal intent or constraints of the other. The scientific challenge, therefore, is not merely whether a given protection succeeds or fails, but why particular perturbation structures persist or collapse under purification, and how these structures are represented inside detection models. Existing evaluations of artist protection largely treat protection and purification as black-box empirical phenomena, measuring success by visual fidelity and output degradation. However, they offer limited insight into **(a)** how protected images are embedded in detection models' latent spaces, **(b)** what internal features signal the presence of perturbations, and **(c)** which signal characteristics (entropy, spatial dispersion, frequency alignment) govern detectability. This work reformulates the protection–purification interaction as an interpretable signal analysis problem spanning three coupled domains: **(1)** Latent representation space (model-internal geometry), **(2)** Activation space (feature-level detection cues), **(3)** Signal space (spatial and spectral structure of perturbations).

### 3.2 Scope and Assumptions

This study is restricted to training-time perturbation defenses for diffusion-based text-to-image generative models. Inference-time adversarial examples and post hoc output filtering are outside the scope of this work, as the objective is to understand how protective signals embedded directly within image data influence learned representations during model training and interact with purification mechanisms prior to deployment. The purification system is modeled as a learned detection–reconstruction operator, rather than a handcrafted signal-processing filter. The defender is assumed to rely solely on image-space statistics and representations learned by an autoencoding architecture, without access to the original clean image or to the optimization objective used during perturbation generation. Detection and purification decisions are therefore made exclusively from the observed input distribution under realistic large-scale data ingestion conditions. No assumption is made that the protection method has access to the internal parameters, gradients, or training dynamics of the downstream diffusion model. Protective perturbations are generated under a model-agnostic threat setting, consistent with real-world artist workflows in which images are protected prior to online publication without knowledge of which models may later ingest the data.

The focus of this study is on interpretability and structural characterization, rather than on proposing a new protection algorithm. The analysis targets how existing perturbation strategies manifest across representation space, activation space, and signal space, and how these manifestations govern detectability and reconstructability under entropy-based purification. Hence, perceptual imperceptibility is assumed to be enforced by the protection methods under study and is therefore not independently re-evaluated. The analysis instead concentrates on model-facing structure, which is the statistical, geometric, and spectral properties of perturbations as observed by learned detection systems.

## 3.3 Research Questions

Based on this formulation, we define three research questions designed to isolate complementary aspects of the protection–detection interaction:

**RQ1 — Representation**: How do detection models represent clean images versus protected images in latent feature space? This question examines whether protection induces distinct geometric structure in the learned embedding space of a purification model, or whether protected samples remain primarily clustered by original visual content. It directly probes whether perturbations create a clearly separable region of the embedding space or remain weakly entangled with clean image structure.

**RQ2 — Detection Mechanism:** Which internal features and activation patterns are used by detection models to identify perturbations? This question targets the operational mechanism of purification networks. Instead of treating detection as an opaque decision, we ask which layers, channels, and spatial responses encode the presence of structured perturbations, and whether different protection methods activate distinct or overlapping internal feature signatures.

**RQ3 — Signal Design and Detectability:** Which perturbation signal characteristics govern detectability and reconstructability under entropy-based purification? This question isolates the image-space properties that control detection success. It explicitly examines how entropy level, spatial non-uniformity, and frequency-band allocation affect whether a perturbation can be reliably reconstructed and removed by a learned purification model.

## 3.4 Unifying Objective

The three research questions defined above jointly reframe the interaction between artist-protection perturbations and purification systems as a multi-domain interpretability problem, rather than as a sequence of isolated empirical observations. Specifically, RQ1 addresses the geometric structure of latent representations, characterizing whether protected and unprotected images occupy separable regions of the feature space of a detection model or remain entangled with semantic content. RQ2 targets the internal detection mechanism itself, probing which layers, channels, and spatial activation patterns encode the presence of perturbations and how these internal signatures differ across protection strategies. RQ3 shifts the focus to the signal-level properties of perturbations, isolating how entropy, spatial dispersion, and frequency allocation govern whether a perturbation can be reliably detected and reconstructed. These three questions establish a unified analytical pipeline that spans representation space, activation space, and signal space. This coupling is essential: latent clustering alone cannot explain what features drive detection; activation analysis alone cannot explain why certain perturbations persist under purification; and signal analysis alone cannot explain how perturbation structure is internalized by learned detection systems. Only by jointly analyzing all three levels can the protection–purification interaction be interpreted as a coherent computational process rather than as a collection of success and failure cases. The unifying objective of this work is therefore to move beyond outcome-based evaluation and toward a structural, interpretable theory of perturbation detectability. Under this view, protection success is not treated as a binary property of a method, but as an emergent consequence of how perturbation structure aligns (or conflicts) with the geometric, spectral, and feature-extraction biases of learned purification models. By grounding artist-protection mechanisms within this multi-level explanatory framework, the study aims to expose the invariants that govern when perturbations remain stable under purification and when they collapse under entropy-based reconstruction. Ultimately, this unification positions XAI not as a post hoc diagnostic tool, but as a central analytical instrument for generative model security, enabling both principled evaluation of existing defenses and informed design of future perturbation strategies.

## 4. Methodology

### 4.1 Overview of Experimental Design

This study employs a controlled, multi-stage experimental pipeline designed to characterize the behavior of perturbation-based image protection methods under detection and purification. The pipeline operates on paired clean and protected images and integrates both internal model instrumentation and model-agnostic signal analysis to capture complementary perspectives on perturbation structure, detectability, and reconstruction. The experimental workflow proceeds in four major stages:

In the first stage, a curated set of visually diverse images is assembled and systematically processed to generate multiple protected variants using contemporary perturbation-based protection methods. These variants, together with their clean counterparts, form the complete dataset used throughout all subsequent analyses. In the second stage, all images are processed through a learned detection–purification model to obtain reconstructed perturbations, detection scores, and intermediate feature representations. In the third stage, internal representations of the purification model are instrumented using direct access to intermediate activations and latent embeddings. This enables measurement of how perturbations propagate through the network hierarchy and how reconstruction behavior emerges across successive feature transformations. In parallel, a model-agnostic analysis track is applied directly in image space, where perturbations are probed using spatial sensitivity experiments, and frequency-domain transforms without access to model internals. In the final stage, structured perturbation patterns with systematically controlled spatial density, masking geometry, and spectral energy distribution are synthesized and evaluated along with the same detection–purification pipeline. This enables direct measurement of how perturbation design parameters influence detectability and reconstruction behavior under identical model conditions.

All experiments are conducted under fixed preprocessing, parameters, and execution settings to isolate the effects of perturbation structure from stochastic variation. The resulting analyses collectively provide a unified empirical foundation for interpreting how protected images are transformed, detected, and reconstructed across both learned representation space and observable signal space. Implementation details, dataset specifications, parameter values, and computational environment are reported separately to ensure full experimental reproducibility.

### 4.2 Datasets and Image Perturbation Pipeline

To ensure controlled and interpretable analysis across protection and detection conditions, we constructed a curated dataset composed of visually diverse images and their systematically generated perturbed variants. The primary dataset consists of nine base images, each selected to represent a distinct artistic style and structural complexity, including painterly, illustrative, and stylized digital forms. This controlled diversity is necessary to prevent stylistic homogeneity from dominating either latent clustering behavior or perturbation detectability. For each base image, four aligned variants were curated under identical preprocessing conditions: **(1)** the original clean image, **(2)** a Glaze-protected version, **(3)** a Nightshade-protected version, and **(4)** a sequentially protected variant produced by applying Nightshade followed by Glaze.

This yields a total of 36 images in the primary white-box analysis dataset. All perturbations were generated using the default protection parameters recommended by the respective tool developers, selected to balance protection strength with minimal visible distortion. No manual post-editing was applied after perturbation generation. All image variants were stored at identical spatial resolution and processed using a unified file format and color space to eliminate confounding effects introduced by compression, rescaling, or color conversion. Each protected image preserves a one-to-one correspondence with its clean source, enabling direct paired analysis throughout all subsequent experiments. All images in the primary dataset were processed through the same detection–purification pipeline to obtain reconstructed perturbations, detection scores, and intermediate feature representations. These outputs form the shared computational substrate for the representational, activation-based, and signal-based analyses described in subsequent sections. In addition to the primary dataset, a separate black-box analysis dataset was constructed to support large-scale spatial sensitivity and frequency-domain experiments. This auxiliary dataset consists of 42 images (Figure 2), including 21 stylized 3D renderings and 21 digital illustrations, selected to span a broad range of geometric structure, shading complexity, and texture density. This dataset is used exclusively for model-agnostic signal characterization and is not included in the white-box detection analysis to avoid cross-contamination between experimental scopes.

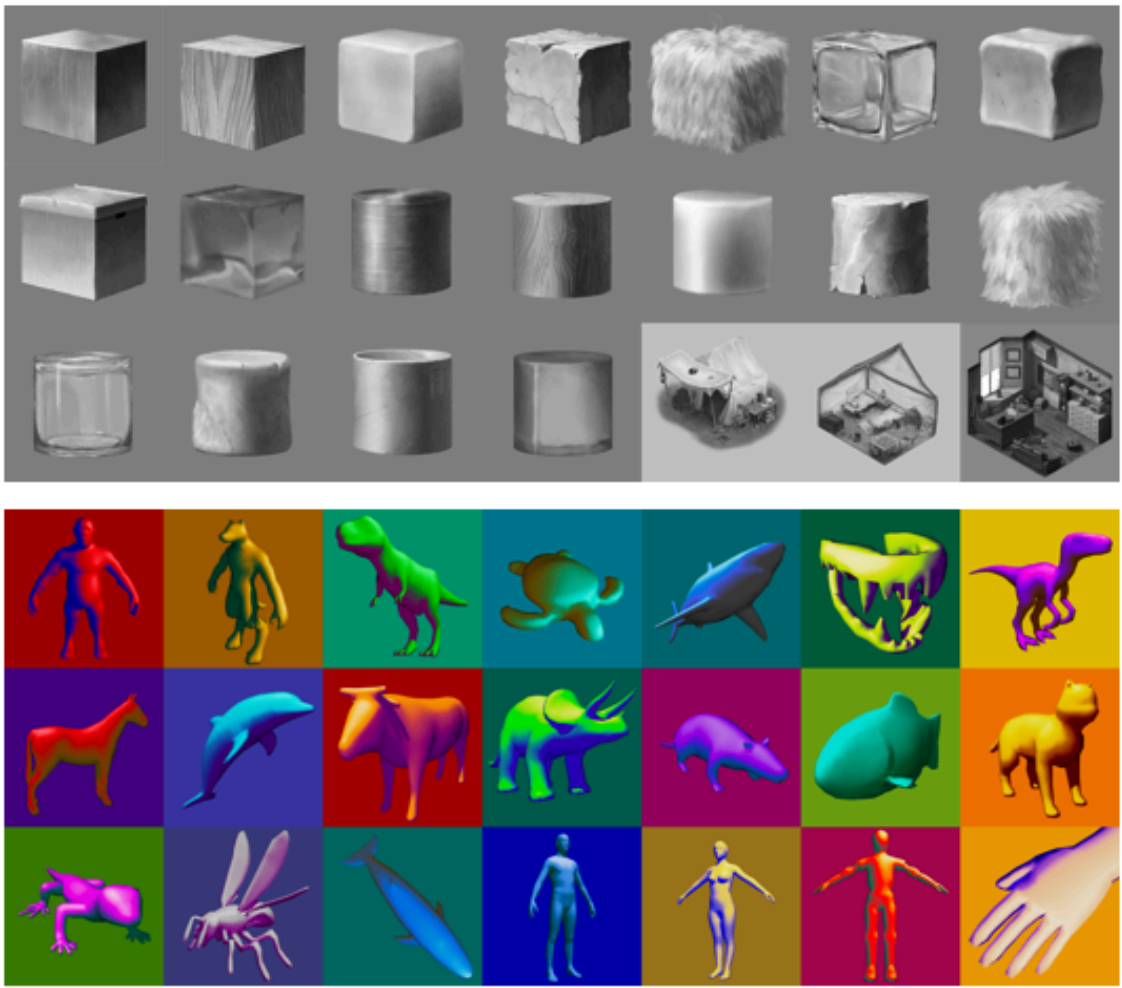

*Figure 2*. Black-box analysis dataset (42 images). Top panel: 21 digital illustrations. Bottom panel: 21 stylized 3D renderings. The dataset spans broad variation in geometry, shading, and texture, and is used exclusively for model-agnostic signal characterization.

For controlled detectability experiments, a set of synthetic perturbation patterns was generated by combining Gaussian noise fields with procedurally defined spatial masks at multiple opacity levels. These perturbations were overlaid onto a fixed, clean base image to produce families of inputs with systematically controlled entropy, spatial dispersion, and frequency structure. All such synthesized variants were processed through the same detection–purification pipeline as the tool-generated perturbations, enabling direct comparability across natural and synthetic perturbation families. Across all datasets, strict correspondence was maintained between clean, protected, reconstructed, and synthesized variants to enable paired statistical comparison and controlled ablation. No images were excluded based on detection outcome, reconstruction quality, or visual appearance, ensuring that all processed inputs contribute uniformly to the empirical analysis.

### 4.3 Protection and Purification Models

This study evaluates contemporary perturbation-based image protection tools in conjunction with a learned detection–purification system. All protection and purification models are treated as fixed, pre-trained components and are not modified or fine-tuned for the purposes of this analysis. This ensures that all observed effects arise solely from the interaction between perturbation structure and model-internal representations rather than from task-specific adaptation.

**Protection Models –** Two state-of-the-art protection frameworks are used to generate training-time perturbations: Glaze and Nightshade. Both methods are applied directly in image space using their publicly released implementations and default parameter configurations recommended by the developers. These default settings are selected to balance perturbation strength with minimal visible distortion and to reflect real-world artist usage. For each clean input image $x$, Glaze produces a protected variant $\tilde{x}_G$, and Nightshade produces a distinct protected variant $\tilde{x}_N$. In addition, a composite protected variant $\tilde{x}_{NG}$ is generated by sequentially applying Nightshade followed by Glaze under identical resolution and color constraints. All perturbations are generated deterministically under fixed random seeds to ensure consistency across experimental runs. No post-processing, denoising, compression, or stylization is applied after perturbation generation. All protected images preserve the original spatial resolution, aspect ratio, and color encoding of their clean counterparts. This guarantees that subsequent detection and reconstruction behavior can be attributed solely to perturbation structure rather than to unintended preprocessing artifacts.

**Purification Model** – Purification, detection, and perturbation reconstruction are performed using LightShed, a learned entropy-based detection–reconstruction architecture. LightShed operates by projecting input images into a latent representation optimized to suppress semantic content while amplifying structured perturbation patterns. The model then reconstructs an estimate of the perturbation, which is used both to compute a detection score and to generate a purified output via subtraction. All images (clean, protected, and synthetically perturbed) are processed using the same frozen LightShed model under identical inference settings. No training, calibration, or threshold tuning is performed on the datasets used in this study. Detection and reconstruction thresholds are inherited directly from the original LightShed configuration to avoid experimental bias. For each input image $x$, LightShed produces three outputs: **(1)** a reconstructed perturbation estimate $\hat{\delta}$, **(2)** a purified image $x_{clean} = x - \hat{\delta}$, and **(3)** a scalar detection confidence score. All three outputs are retained for downstream representational, activation-based, and signal-level analyses.

**Integration into the Experimental Pipeline** – Protection and purification models are integrated into the experimental pipeline as deterministic, feed-forward operators. Each image variant is processed independently without batch-dependent coupling or adaptive feedback. This ensures that inter-image statistical dependencies do not confound latent clustering, activation analysis, or detectability measurements. All model executions are performed under fixed hardware, preprocessing, and numerical precision settings. The resulting perturbation estimates, detection scores, and intermediate activations serve as the shared computational substrate for all subsequent analyses described in Sections 4.4–4.8.

## 4.4 Latent-Space Clustering and Representational Analysis
To examine how protected and unprotected images are embedded within the internal representation space of the purification model, we perform a latent-space clustering analysis based on intermediate feature embeddings extracted from the detection–purification network. This analysis is designed to characterize the geometric organization of clean, protected, and reconstructed images within a common feature space. For each input image, feature vectors are extracted from a fixed intermediate layer of the purification model, selected to balance representational abstraction with spatial resolution. These model-produced embeddings serve as high-dimensional descriptors capturing the model's internal encoding of both semantic content and perturbation structure. Feature extraction is performed under identical inference settings for all image variants to ensure comparability across conditions. Because the dimensionality of these embeddings precludes direct visualization, nonlinear dimensionality reduction is applied to project the feature vectors into a two-dimensional feature space. We employ t-distributed stochastic neighbor embedding (t-SNE) with a fixed perplexity and random seed to preserve local neighborhood structure while enabling consistent comparisons across experimental runs. All projections are computed jointly over clean, protected, and reconstructed image embeddings to avoid projection bias. Clustering behavior is evaluated by inspecting the relative spatial organization of clean images, protected variants, and purified outputs within the reduced projection. Rather than enforcing predefined class labels, the analysis is performed in an unsupervised manner to reveal whether perturbation structure induces separable representational modes or remains embedded within semantic content clusters. Inter-point distances and cluster dispersion metrics are computed to quantify degrees of separation and overlap. To assess reconstruction effects, embeddings of purified images are embedded alongside their clean and protected counterparts. This allows direct measurement of whether the purification process restores representations toward their original clean feature space or produces novel hybrid structures. All comparisons are performed on paired image triplets to preserve one-to-one correspondence across conditions.

To verify that observed clustering behavior is not an artifact of a single projection configuration, all t-SNE analyses are repeated under multiple perplexity settings and embedding initializations. The stability of cluster topology across runs is used as a consistency check on representational structure. This latent-space analysis establishes a geometric characterization of how perturbation structure alters internal image representations and how reconstruction modifies these representations under the purification model. The resulting embeddings form the basis for the representational-level findings described in Section 5.

## 4.5 Layer-Wise Feature Activation Analysis

To characterize how perturbation structure is encoded across the internal feature hierarchy of the purification model, we perform a systematic layer-wise feature activation analysis using direct instrumentation of intermediate network layers. This analysis quantifies how protected and unprotected images differentially excite internal feature responses and how these responses evolve across depth. The purification model is instrumented with forward hooks at a fixed set of convolutional and bottleneck layers spanning early, mid-level, and deep feature representations (Foerster et al., 2025). These layers are selected to capture feature responses ranging from low-level texture and edge structure to higher-level, spatially aggregated representations. For each input image, full activation tensors are recorded under identical inference conditions. Activation responses are analyzed using both global and spatially resolved statistics. Channel-wise activation magnitudes are aggregated to quantify overall response intensity, while spatial activation maps are retained to evaluate localized sensitivity patterns. For each layer, activation distributions are computed independently for clean images, protected variants, and purified outputs for direct paired comparison across conditions. To isolate perturbation-specific responses, we compute activation difference maps by subtracting clean-image activations from corresponding protected-image activations at identical spatial locations. These difference maps serve as internal saliency representations of perturbation-sensitive features and enable identification of channels that function as high-sensitivity detectors (Selvaraju et al., 2017; Sundararajan et al., 2017). Activation sparsity, peak-to-mean ratios, and spatial concentration metrics are used to quantify the structure of these internal responses. To assess reconstruction effects, the same activation statistics are computed for purified images and compared against both clean and protected baselines, which enables measurement of whether purification suppresses perturbation-driven activations uniformly across layers or selectively at specific depths within the network hierarchy. All activation analyses are performed on paired image triplets to preserve one-to-one correspondence across clean, protected, and reconstructed conditions. Results are aggregated across images within each perturbation class to assess consistency of internal detection patterns while avoiding dependence on any single input. This layer-wise analysis provides a feature-space characterization of how perturbations are encoded, amplified, or suppressed across the internal computation of the purification model. The resulting activation statistics form the basis for the internal mechanism findings described in Section 5.

## 4.6 Detectability Experiments: Mask–Noise Compositions and Entropy Analysis

We design a controlled family of synthetic perturbation patterns with precisely regulated entropy, spatial dispersion, and amplitude distribution to systematically evaluate how perturbation structure influences detectability and reconstruction behavior. These perturbations are used to probe the sensitivity limits of the detection–purification model under tightly controlled signal conditions that isolate structural factors from semantic image content. All synthetic perturbations are generated by combining Gaussian noise fields with procedurally defined binary and soft spatial masks spanning global, regional, band-limited, and sparse structural patterns.

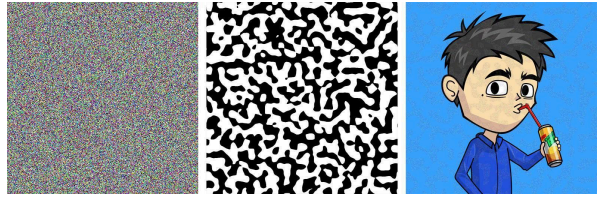

*Figure 3:* *Synthetic perturbation generation via noise–mask composition. From left to right: Gaussian noise field, procedural spatial mask, and the resulting masked perturbation applied to the clean reference image.*

Figure 3 illustrates the noise–mask composition process, showing the Gaussian noise field, the corresponding procedural mask, and the resulting masked perturbation applied to the clean reference image. A step-by-step breakdown of this process is provided in Figure 4.

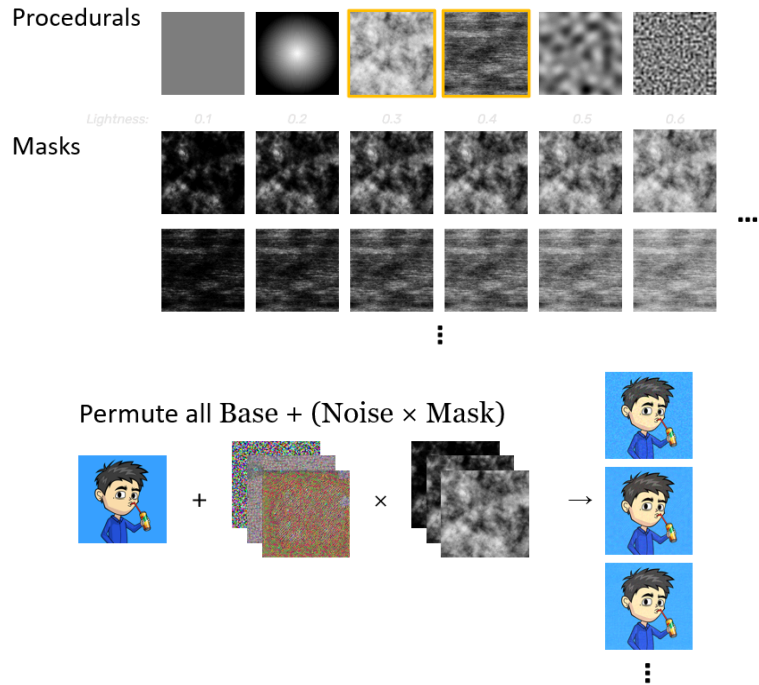

*Figure 4:* *Visualized process shown in Algorithm 1. Top Panel: Procedural textures and derived lightness-controlled masks – A set of procedural noise patterns is transformed into multiple masks by applying gamma correction at different lightness levels. Lightness is defined as the mean pixel intensity of each mask. This produces a family of masks spanning low to high average opacity while preserving the same spatial structure. Bottom Panel: Noise–mask compositing pipeline – Each base image is combined with every noise sample and every lightness-controlled mask using elementwise multiplication and additive compositing. The resulting perturbed images are clipped to the valid intensity range and exported. These outputs will then be subsequently processed by LightShed to reconstruct the injected poisoning noise.*

The full noise–mask permutation procedure used to synthesize these controlled perturbation families is showcased in Algorithm 1. All noise–mask compositions are applied to a fixed clean reference image $b$ to ensure that variation across perturbation families arises solely from controlled signal structure. The opacity parameter $a$ is systematically varied across experiments to generate perturbations with progressively increasing Shannon entropy. For each geometry, noise intensity is scaled across multiple opacity levels to generate perturbations with progressively increasing Shannon entropy. This design enables independent control over perturbation density and signal randomness while maintaining identical base image content across all variants. Each masked-noise perturbation is overlaid additively onto a fixed clean reference image to produce a family of inputs with strictly controlled structural variation. The resulting images are then processed through the same detection–purification pipeline used for tool-generated perturbations, yielding detection confidence scores and reconstructed perturbation estimates under identical inference conditions.

**Algorithm 1** Noise and Mask Permutation

    Given base image b, masks M, and noises N         ▷ Pixels in b are in [0, 255]
      **for** m ∈ M **do**         ▷ Pixels in m are in [0, 1]
        **for** n ∈ N **do**         ▷ Pixels in n are in [0, 255]
          m ← m ∗ 0.15         ▷ Master opacity 15%
                (Master opacity $\alpha$ controlling perturbation entropy)

          c ← b + n ⊙ m
          c ← Clip(c, 0, 255)
          Export c
      **end for**
    **end for**

For each perturbation family, detection sensitivity is quantified as a function of both mask geometry and noise entropy level. Reconstruction fidelity is evaluated by comparing reconstructed perturbations against their ground-truth synthetic sources using pixel-wise error statistics and structural similarity measures. These evaluations provide complementary views of detectability: one through decision confidence and one through reconstruction accuracy. To assess robustness to weak perturbation regimes, a subset of experiments is conducted at extremely low noise amplitudes near perceptual invisibility thresholds. This regime tests whether detectability arises from aggregate structural biases or requires high-energy signal components to trigger reconstruction. All synthetic experiments are paired with equivalent tool-generated perturbations applied to the same base image to enable direct cross-family comparisons between naturalistic and procedurally controlled perturbation structures. This pairing ensures that conclusions regarding entropy and spatial structure generalize beyond any single protection mechanism.

### 4.7 Spatial Sensitivity Analysis via Occlusion-Based Probing

To characterize how perturbations depend on local image content, we perform an occlusion-based spatial sensitivity analysis that operates purely in image space. The analysis compares a perturbed image to systematically occluded versions of its corresponding clean image, yielding a model-agnostic measure of how strongly perturbation structure is tied to specific spatial regions. For each clean–protected image pair, let *b* denote the clean baseline image and *p* the corresponding perturbed image. The complete occlusion-based spatial sensitivity mapping procedure is formally specified in Algorithm 2.

**Algorithm 2** Occlusion Sensitivity Mapping

Given a clean image b ∈ [0,1], a perturbed image p ∈ [0,1], an occlusion window size w, and stride s
H ← zeros_like(b)         ▷ Heatmap, initialized to 0
**for** y ∈ {0, …, **H**(b) − w, step s} **do**
      for x ∈ {0, …, W(b) − w, step s} do
          b_occ ← b         ▷ Copy clean image
          b_occ[y : y+w, x : x+w] ← 0         ▷ Apply occlusion patch
                (Replace w×w region with zeros)

          score ← | b_occ − p |         ▷ Pixelwise absolute difference
          score ← **Mean**(score)         ▷ Global scalar difference
          **H**[y : y+w, x : x+w] ← score         ▷ Fill heatmap region with score
      **end for**
    **end for**

    **Export H**         ▷ Sensitivity map for perturbation p

A sliding occlusion window is applied across *b* using a fixed window size and stride. At each spatial location, the pixels within the window are replaced with a neutral zero baseline, producing an occluded baseline image $b_{occ}$. This operation removes local structure in the clean image while leaving the perturbed counterpart *p* unchanged. To quantify the sensitivity at each occlusion position, we compute the mean pixel-wise absolute difference $|b_{occ} - p|$ over the full image domain. The resulting scalar value is assigned to the spatial location of the occlusion window, yielding a dense grid of sensitivity scores. Collectively, these scores form a spatial sensitivity map, where higher values indicate regions in which the perturbation signal is more dependent on the underlying clean-image content that has been zeroed out. This procedure is applied independently to each protection method and to each image in the dataset. Within each image, sensitivity maps are normalized to facilitate comparison across images with differing overall intensity ranges and perturbation magnitudes. The normalized maps are then aggregated across images within each perturbation family to identify consistent spatial patterns in how perturbations relate to local structure. This occlusion-based analysis provides a model-agnostic spatial decomposition of perturbation dependence on local image content, without relying on internal gradients, detection scores, or architectural assumptions. The resulting sensitivity distributions form the basis for the spatial structure results discussed in Section 5.

## 4.8 Fourier and Frequency-Domain Characterization of Perturbations

To analyze the spectral structure of perturbation signals independently of spatial localization and model internals, we perform a frequency-domain characterization using two-dimensional Fourier analysis. This analysis reveals how perturbations redistribute energy across spatial frequencies and enables direct comparison between clean images and their Glaze, Nightshade, and composite-protected variants. Frequency-domain difference maps are obtained as signed differences between the perturbed and clean log-magnitude spectra, directly highlighting frequency bands in which perturbation structure increases or suppresses spectral energy relative to the original image content. The complete Fourier-based perturbation fingerprint extraction pipeline is formally specified in Algorithm 3.

---

**Algorithm 3** Fourier-Based Perturbation Fingerprint Pipeline (parent algorithm)

---

Given a clean image b ∈ [0,1], a perturbed image p ∈ [0,1]

**# 1. Convert to grayscale**
$b_{gray}$ ← MeanChannel(b)                  ▷ Convert to grayscale (H×W)
$p_{gray}$ ← MeanChannel(p)

**# 2. Compute log-magnitude FFT spectra (Algorithm 3A)**
$M_b$ ← FFT_LogMagnitude($b_{gray}$)         ▷ Log |FFT| of clean image
$M_p$ ← FFT_LogMagnitude($p_{gray}$)         ▷ Log |FFT| of perturbed image

**# 3. Signed frequency-domain difference**
ΔF ← $M_p$ – $M_b$                           ▷ Positive = gain in energy, negative = loss

**# 4. Radial frequency profiles (Algorithm 3B)**
($R_b$, $S_b$) ← RadialProfile($M_b$)   ▷ Clean radial spectrum curve (radii + averaged magnitudes)
($R_p$, $S_p$) ← RadialProfile($M_p$)   ▷ Perturbed radial spectrum curve

(Note: Radii are identical for both ▢ $R_b$ = $R_p$ = R)

**Export**:
$M_b$ ▷ FFT magnitude spectrum (clean)
$M_p$ ▷ Perturbed FFT log-magnitude spectrum
**ΔF** ▷ Signed FFT difference map
R ▷ Shared radius vector
$S_b$ ▷ Clean radial magnitude profile
$S_p$ ▷ Perturbed radial magnitude profile

For each clean–protected image pair, both images are first converted to grayscale through channel averaging so that spectral measurements reflect structural variation rather than color-channel imbalance. The complete Fourier-based perturbation fingerprint extraction pipeline is formally specified in Algorithm 3. The two-dimensional discrete Fourier transform (2D-DFT), implemented via the Fast Fourier Transform, is then applied to each grayscale image, and log-magnitude spectra are computed with the zero-frequency component (Direct Current component) shifted to the center of the spectrum for visual and analytical consistency. The detailed FFT and log-magnitude spectrum computation procedure is given in Algorithm 3A.

**Algorithm 3A** Log-Magnitude FFT Computation (Grayscale)

Given grayscale image g ∈ [0,1], size H×W
F ← FFT2(g)   ▷ 2D Fast Fourier Transform
F ← FFTShift(F)   ▷ Shift DC component to the center
M ← log(1 + |F|)   ▷ Log-magnitude for dynamic range compression
Return M   ▷ Spectral energy representation

To obtain a compact, direction-invariant summary of spectral redistribution, radially averaged frequency profiles are computed by averaging spectral magnitude over concentric rings of constant radius in the frequency plane. The radial frequency profile extraction procedure used for all spectral summarization is specified in Algorithm 3B.

**Algorithm 3B** Radial Frequency Profile (Grayscale)

Given magnitude spectrum $M \in \mathbb{R}^{H \times W}$

(cx, cy) ← (W/2, H/2)   ▷ Spectrum center
r(x,y) ← **round**( sqrt((x−cx)² + (y−cy)²) )   ▷ Radius for each pixel

Let r_max ← max radius in image
Initialize arrays:
    **R** ← {0,1,...,r_max}   ▷ Radii
    **S** ← zeros(r_max + 1)   ▷ Radial averages

for k ∈ {0, …, r_max} do
    mask ← {(x,y) | r(x,y) = k}
    S[k] ← Mean(M[mask])   ▷ Average magnitude at radius k
end for

Return (R, S)   ▷ Radius vector and radial magnitude profile

This produces a one-dimensional radial profile for each perturbation variant, explicitly encoding how spectral energy is redistributed from low to high spatial frequencies. These profiles support quantitative comparison of spectral concentration, high-frequency amplification, and bandwidth occupancy across perturbation types. The same Fourier analysis pipeline is applied consistently across all image groups and protection variants under fixed spatial resolution and identical normalization conventions. All spectral measurements are computed on paired clean–protected images to ensure that observed frequency-domain effects arise solely from perturbation structure rather than content variation. The resulting frequency-domain representations and radial profiles form the basis for the spectral-structure findings reported in Section 5.

### 4.9 Implementation Details and Reproducibility

All experiments are implemented in a unified Python-based pipeline to ensure consistency across protection, detection, and analysis stages. All protection methods are executed using the official public implementations of Glaze and Nightshade under their recommended default configurations. The detection–purification model is evaluated strictly in inference-only mode with frozen weights throughout all experiments. No fine-tuning, post-hoc calibration, or threshold adjustment is applied. To ensure strict experimental control, all stochastic components (including noise generation, mask sampling, and dimensionality-reduction initialization) are executed with fixed random seeds. All images are processed at fixed spatial resolution and under identical normalization conventions across all experiments. To support full reproducibility, the complete experimental configuration (including parameters governing noise amplitude, mask geometry, occlusion window size and stride, FFT normalization, and radial binning resolution) is fixed across all reported results. The procedures governing synthetic perturbation synthesis, occlusion-based probing, and frequency-domain analysis are formally specified in Algorithms 1–3B. This ensures that all reported findings are attributable to perturbation structure and model behavior rather than to uncontrolled preprocessing, parameter drift, or stochastic variation.

## 5. Results and Analysis
### 5.1 Representational Separation in Feature Space

This subsection examines how clean and protected images are organized within the internal feature space of the detection–purification model following the feature extraction and nonlinear projection procedure described in Section 4.4. The analysis focuses on whether different protection mechanisms induce systematic, structured displacements in feature space relative to the clean semantic representations, and how these displacements relate to underlying image content rather than purely to the protection method itself. Figure 5 presents a manually annotated two-dimensional t-SNE visualization of clean, Glaze-protected, Nightshade-protected, and sequentially protected image representations.

Across the evaluated image set, images cluster primarily according to their base visual content. In other words, clean and protected variants of the same image remain grouped within the same dominant cluster, indicating that the purification model preserves strong semantic and stylistic organization even under perturbation. This behavior is expected, as Glaze and Nightshade are designed to introduce perceptually subtle perturbations that do not visually alter high-level content. These results confirm that protection methods do not induce global representational drift in feature space. Within each base-image cluster, however, consistent protection-specific substructure emerges. Clean images and Nightshade-protected images form distinguishable subclusters, while Glaze-protected images and sequentially protected (Nightshade-glaze) images consistently overlap within a shared subregion. This pattern reflects the empirical dominance of the second applied perturbation, whereby the final protection stage largely determines the visible and latent characteristics of the resulting signal. As a result, sequentially protected images inherit the feature-space behavior of the method applied last, consistent with observations from both spatial and spectral analyses. Importantly, this subclustering behavior appears consistently across multiple artistic styles and scene types, including digital paintings, 3D renders, and illustrated environments. This confirms that the observed feature-space organization arises from protection structure operating within each image's semantic embedding, rather than from differences in artistic genre or scene content. Together, these results establish that Glaze and Nightshade operate as structured, low-magnitude feature perturbations that remain tightly coupled to the original image manifold, while still introducing detectable method-specific substructure within that feature space.

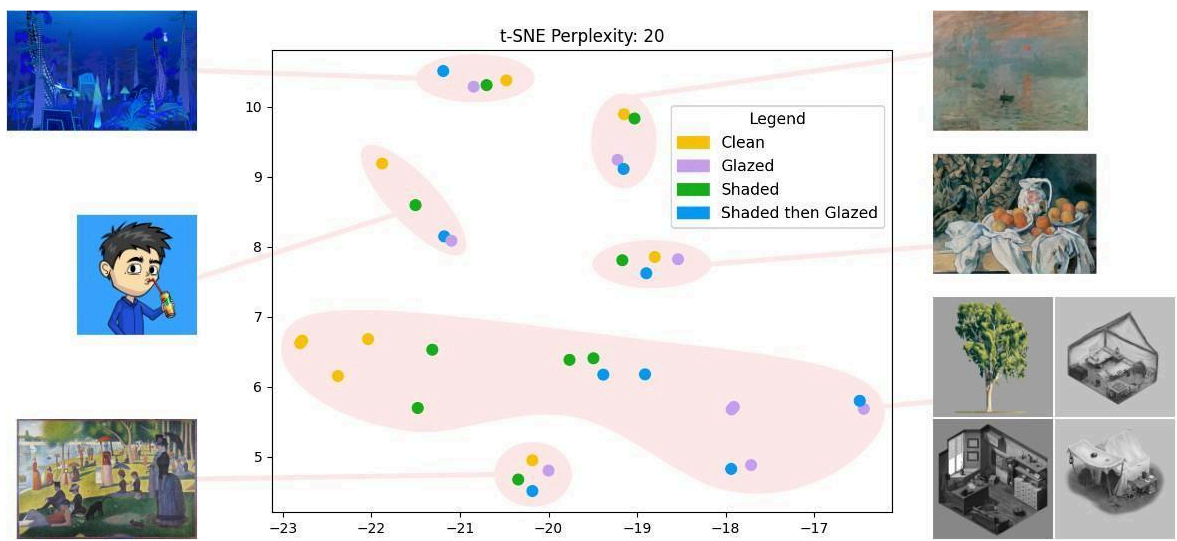

*Figure 5: Manually annotated t-SNE visualization of image latent representations under different protection conditions. Each cluster corresponds to a shared base image, with clean, Glaze-protected, Nightshade-protected, and sequentially protected variants color-coded as indicated. Representations cluster primarily by visual content rather than protection method, while protection-specific subclustering emerges within each content group, indicating structured perturbation effects without global semantic drift.*

## 5.2 Internal Activation Behavior Under Image Protection

This subsection analyzes how image protection mechanisms modulate the internal activation dynamics of the detection–purification model across multiple feature-processing stages. The objective is to characterize how perturbation energy is distributed, amplified, or suppressed as protected images propagate through the network's hierarchical representation. Across the evaluated layers shown in Figure 6, protected images consistently induce higher-magnitude and more spatially concentrated activations relative to their clean counterparts. This effect is most pronounced in the second and, in particular, the third encoder layers, which exhibit strong, spatially structured activation patterns in response to protected inputs. In these layers, multiple channels become highly responsive across broad image regions under protection, whereas clean images produce smoother, lower-variance activation profiles that reflect underlying scene structure rather than perturbation structure. In contrast, deeper layers exhibit markedly reduced activation sensitivity to protection, with many channels in the fourth and fifth layers showing little to no measurable response. This attenuation indicates that perturbation-related signal energy is detected and amplified primarily at intermediate feature-processing stages rather than at the deepest representational levels. These observations demonstrate that Glaze and Nightshade introduce structured, mid-level signal components that are preferentially captured by a small subset of encoder layers, providing a mechanistic explanation for why LightShed's detection and reconstruction capabilities are dominated by its intermediate feature representations.

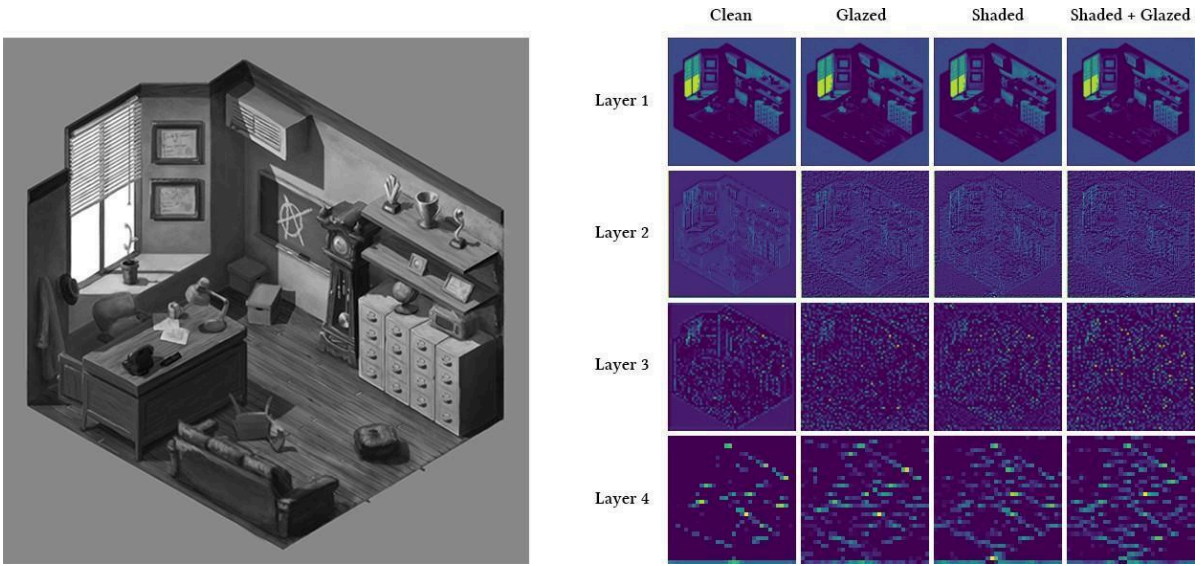

*Figure 6:* *Left: Clean input image of an isometric room. Right: Layer-wise activations of a selected feature channel across LightShed's encoder for the clean image and each poisoning technique. Glaze and sequential (Shaded-Glazed) protection induce the strongest activation noise in the intermediate layers.*

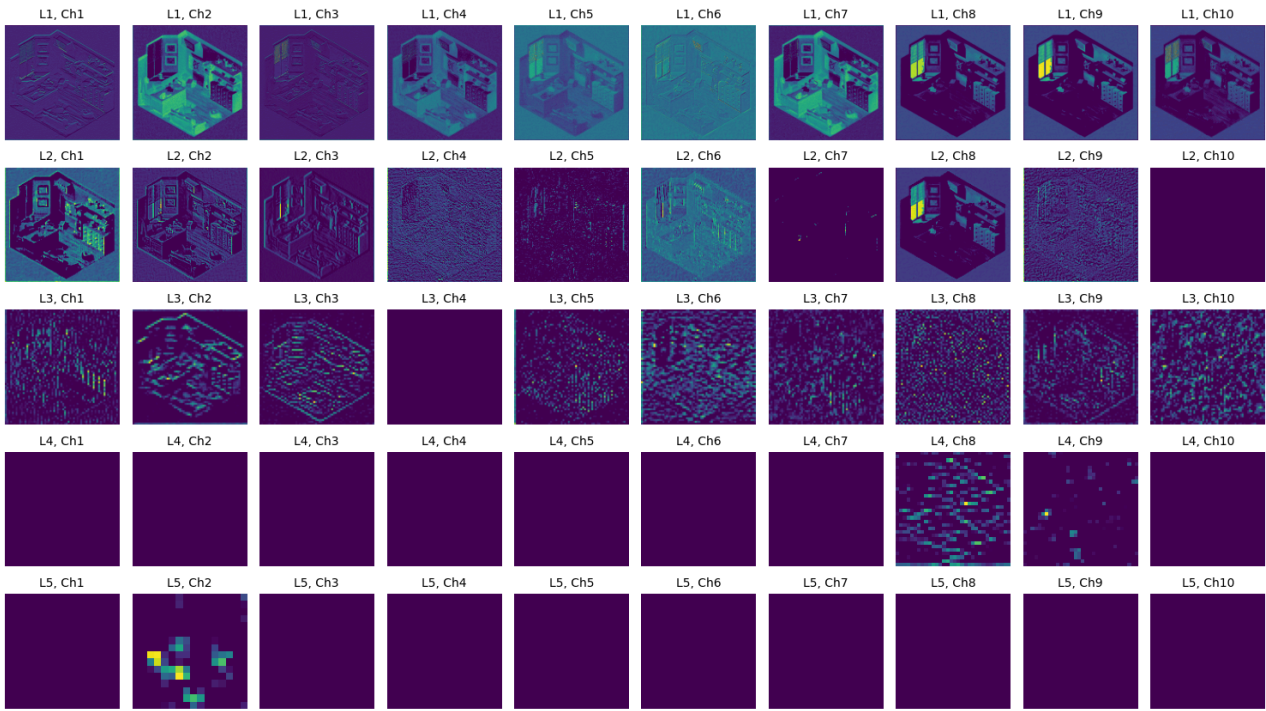

*Figure 7:* *Layer-wise activation maps for a representative Shaded–Glazed image across the first five encoder layers (L1–L5) of the "isometric room" scene showcased in Figure 6. Ten channels are shown per layer. Early layers (L1–L2) preserve strong spatial structure aligned with the input image. Mid-level layers (L3) exhibit sparse, high-variance perturbation-driven activations. Deep layers (L4–L5) show near-zero responses across most channels, indicating strong attenuation of protection signals at higher representational depths.*

Nightshade-protected images exhibit the strongest activation amplification, with high-energy responses persisting across successive layers, indicating that semantic poisoning introduces perturbation structures that remain salient under repeated nonlinear transformation. Glaze-protected images, by contrast, produce moderate activation elevation, with excitation patterns that attenuate more rapidly with depth. Sequentially protected images follow the activation signature of the final applied protection stage, reflecting perturbation dominance at the representational level. Activation difference maps further reveal that protection-sensitive responses are not uniformly distributed across the feature hierarchy (Figure 7). Instead, a small subset of channels exhibits disproportionately high responses, acting as the dominant perturbation-sensitive feature channels at the mid-level of the network. These channels show consistent excitation across images and styles, indicating that perturbation encoding is mediated by stable internal mechanisms rather than image-specific coincidences. As shown in Figure 8, for Layer 3, some channels exposed to protected images exhibit incoherent, noise-like activation patterns, while others retain key geometric structure from the underlying image. In contrast, the corresponding activations for clean images remain smooth and spatially structured. This selective transition from coherent to noisy activation across channels indicates that protection mechanisms selectively disrupt specific feature pathways rather than uniformly degrading representation quality.

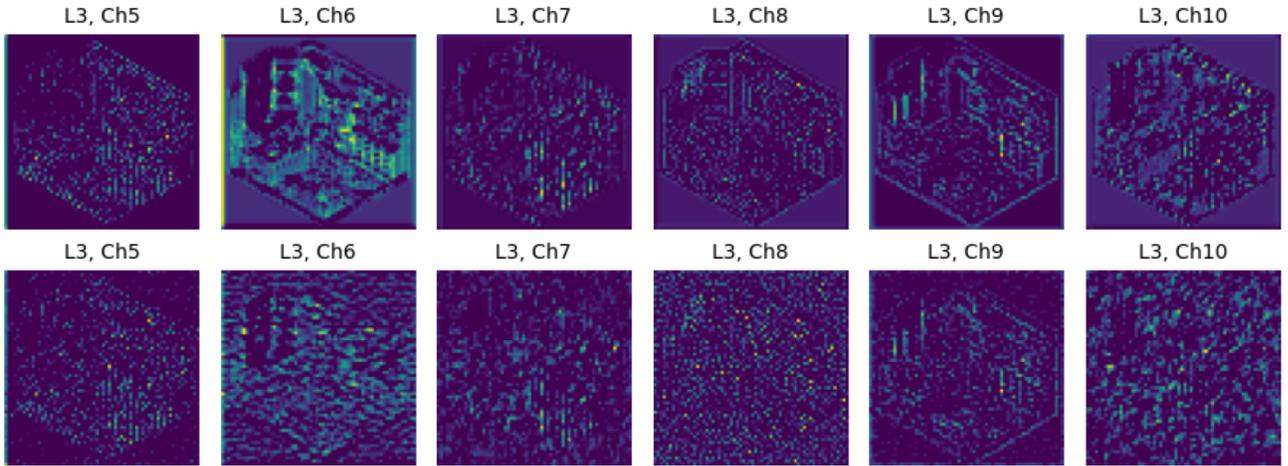

*Figure 8:* Channel-wise activation comparison for six representative channels in Layer 3 between a clean image (top row) and a sequentially protected Shaded–Glazed image (bottom row). Clean activations retain coherent geometric structure, whereas protected activations exhibit channel-specific transitions to noise-like or edge-amplified patterns. This illustrates that perturbation encoding concentration is focused on a sparse subset of mid-level feature channels rather than a uniform distribution across the layer.

As network depth increases, the spatial structure of perturbation-driven activations becomes increasingly diffuse, indicating progressive entanglement of perturbation signals with higher-level representational abstractions. However, detectable activation asymmetries persist even in deeper layers, demonstrating that protection signals are not fully suppressed by semantic abstraction. Collectively, these results demonstrate that image protection mechanisms operate not only through pixel-level distortions but also through structured internal modulation of feature-channel responses that propagate across the network hierarchy. This feature-level modulation provides a mechanistic link between the content-aligned embedding-space organization observed in Section 5.1 and the detectability, spatial sensitivity, and spectral behaviors examined in the subsequent analyses.

### 5.3 Detectability Trends Under Controlled Entropy Variation
#### 5.3.1 Tool-Based Detectability (Glaze, Nightshade, Sequential Protection)
This subsection reports the empirical detectability behavior of real-world image protection tools under the learned detection–purification model. Detection responses are evaluated for images protected using Glaze, Nightshade, and sequential (Shaded-Glaze) perturbations under identical inference conditions across all samples. Aggregate reconstruction entropy and detection rates are summarized in Table 1.

| Mask | Entropy | Detect | Noise | Entropy | Detect | L | Entropy | Detect |
|---|---|---|---|---|---|---|---|---|
| Uniform | 1.231 | 60.4% | Gauss | 2.418 | 77.1% | 0.1 | 0.048 | 30.6% |
| Radial Gradient | 1.311 | 60.4% | Gauss-2x | 0.164 | 41.7% | 0.2 | 0.303 | 50.0% |
| Clouds2 | 1.177 | 64.6% | Gauss-4x | 0.002 | 0% | 0.3 | 0.700 | 61.1% |
| Directional | 1.152 | 62.5% | Glazed | 0.908 | 50.0% | 0.4 | 1.166 | 63.9% |
| Hi-Freq Perlin | 1.143 | 52.1% | Shaded | 1.271 | 93.8% | 0.5 | 1.416 | 72.2% |
| Low-Freq Perlin | 1.208 | 62.5% | S+G | 2.460 | 100% | 0.6 | 1.688 | 72.2% |
| | | | | | | 0.7 | 2.049 | 66.7% |
| | | | | | | 0.8 | 2.261 | 66.7% |

*Table 1: Average Shannon entropy of reconstructed perturbations and corresponding LightShed detection rates across all synthetic perturbation families. Results are reported as a function of spatial mask pattern (left), noise type (middle), and mask lightness (right). Lower entropy values indicate reduced reconstructability and are associated with lower detection probability.*

Across the evaluated image set, detection responses exhibit clear method-dependent variation. Nightshade-protected images consistently produce strong detection responses (average reconstruction entropy 1.271, detection rate 93.8%), indicating that semantic poisoning introduces perturbation structures that remain readily detectable by the purification model. Glaze-protected images, in contrast, yield systematically lower detection confidence (entropy 0.908, detection rate 50.0%), reflecting the design objective of stylistic camouflage to minimize detectable footprint while altering downstream model behavior. Sequentially protected images exhibit the highest measured reconstruction entropy (2.460) and a 100% detection rate in Table 1. This indicates that, in this experimental configuration, sequential application amplifies rather than suppresses the detectable perturbation structure under the purification model. The resulting detection behavior is non-additive, reinforcing the dominant poisoned signal rather than attenuating it. This outcome is consistent with the feature-space subclustering observed in Section 5.1 and the activation-level dominance reported in Section 5.2, where the final applied perturbation governs internal representation. Detection responses remain consistent across stylistic categories and image content, confirming that tool-based detectability is governed primarily by perturbation structure rather than semantic or artistic variation.

### 5.3.2 Synthetic Mask–Noise Detectability

To isolate the fundamental signal-level factors governing detectability, we evaluate procedurally generated perturbations synthesized from controlled combinations of spatial masks, noise fields, and opacity levels, as defined in Algorithm 1. A total of 288 synthetic perturbations spanning all mask, noise, and lightness combinations are processed through the LightShed detection–purification model. For each input, LightShed reconstructs the estimated perturbation and computes the Shannon entropy of the reconstruction. As shown in Figure 9, an image is classified as poisoned when this entropy exceeds the default detection threshold of approximately 0.07. Average reconstruction entropy and detection rate are reported as a function of mask pattern, noise type, and mask lightness in the analyzed statistics shown in Table 1, along with the accompanying entropy visualization (Figure 10).

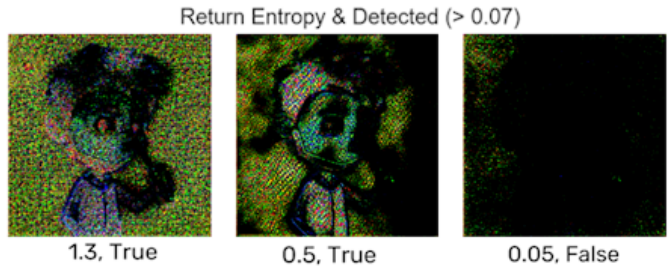

*Figure 9: LightShed poison reconstructions of synthetic mask-noise, annotated with the corresponding Shannon entropy and binary detection outcome using a fixed threshold of 0.07.*

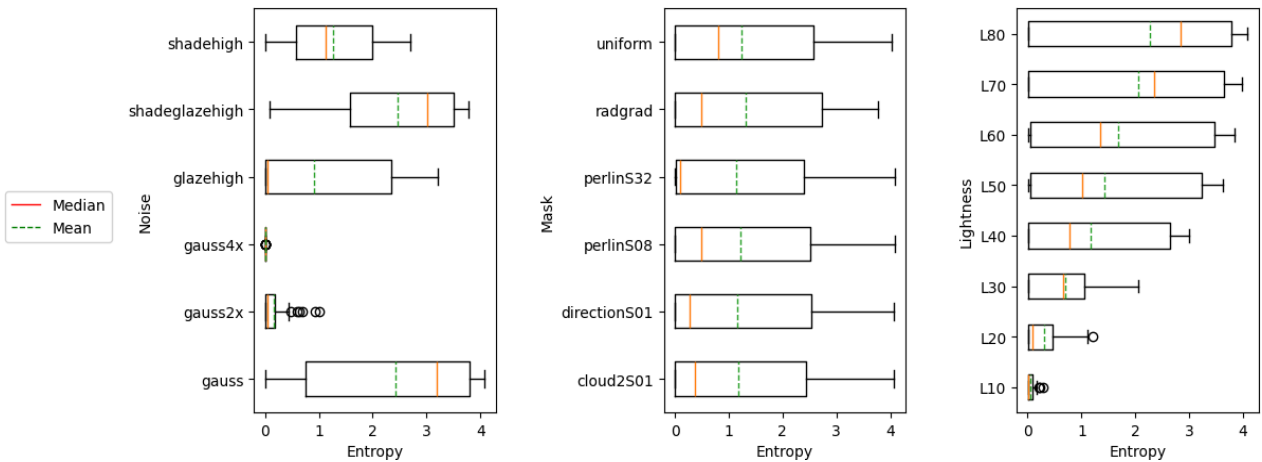

*Figure 10:* *Distribution of Shannon entropy for LightShed-reconstructed synthetic perturbations as a function of noise type (left), spatial mask geometry (middle), and mask lightness (right). Each boxplot summarizes entropy across all generated samples for the corresponding condition. Median (solid red) and mean (dashed green) are indicated. Noise structure exerts the strongest influence on reconstruction entropy, with Glaze- and Nightshade-derived noise producing the highest entropy and upscaled Gaussian noise producing near-zero entropy. Mask geometry induces weaker variation, while increasing mask lightness produces a monotonic rise in reconstruction entropy.*

Across mask geometries, average reconstruction entropy remains relatively stable (approximately 1.1–1.32), indicating that procedural spatial structure alone induces comparable levels of reconstructability and detectability across mask families. Low-frequency Perlin perturbations yield higher entropy than high-frequency Perlin noise, with the latter producing the lowest detection rate (approximately 52%) among all mask types. All other procedural masks exceed 60% detection on average. This confirms that, at the mask level, most spatial patterns generate broadly similar detection behavior under equivalent entropy. Noise structure exerts the dominant influence on detectability. Upscaled Gaussian noise (Gauss-4x) produces the lowest average entropy (approximately 0.002) and is never detected, whereas Glaze-derived noise and sequential Nightshade-glaze noise produce the highest entropy values (approximately 2.4) and are detected in nearly all cases.

This disparity reflects a strong model–perturbation alignment effect: LightShed is explicitly trained to reconstruct and detect Glaze- and Nightshade-style perturbations, whereas upscaled Gaussian noise lies outside its learned perturbation distribution, making it both difficult to reconstruct and unlikely to be detected. Mask lightness further modulates reconstructability and detectability. Average reconstruction entropy increases monotonically with lightness, from $L = 0.1$ (entropy $\approx 0.05$) to $L = 0.8$ (entropy $\approx 2.26$), indicating that stronger perturbation amplitude directly amplifies the reconstructed signal. However, the detection rate peaks at intermediate lightness values ($L \approx 0.5$–$0.6$) and declines at higher opacity. This non-monotonic behavior suggests that when excessive noise is present, the perturbation becomes increasingly difficult to distinguish from legitimate artistic texture, reducing detector confidence despite the higher entropy. These results demonstrate that entropy magnitude, noise distribution, and spatial deployment act as separable but interacting controls over detectability. While real-world protection tools embed semantic intent into their perturbation structure, this synthetic analysis reveals the fundamental signal-level constraints that govern when perturbations become reconstructable and detectable under learned purification models.

## 5.4 Spatial Dependence of Perturbation Structure

To characterize how protection mechanisms depend on the spatial structure of the underlying image, we analyze occlusion-based spatial sensitivity maps computed from clean–protected image pairs. This analysis provides a model-agnostic measure of spatial coupling between perturbation structure and local image content by quantifying how localized occlusions of the clean image modulate the clean–protected difference. Regions of high sensitivity indicate strong structural dependence on the underlying image content, whereas low-sensitivity regions correspond to perturbations that are spatially diffused or weakly coupled to local structure. Across all evaluated examples, including stylized 3D models and material studies, a consistent global trend emerges: all perturbation types (Glaze, Nightshade, and their sequential combination) remain spatially anchored to the clean image. None of the protection methods behaves as spatially independent noise fields. Instead, their influence is systematically modulated by the geometry, shading, and structurally salient regions of the original artwork. For articulated 3D objects such as the stylized cow, wasp, and human models in Figure 11, occlusion sensitivity maps reveal method-dependent spatial signatures. Glaze exhibits a more object-localized dependence, with the strongest responses concentrated near high-curvature regions, sharp edges, and areas of strong geometric contrast. In foreground-dominated scenes, this produces elevated sensitivity over object features and comparatively suppressed background responses, indicating tight coupling to localized surface structure.

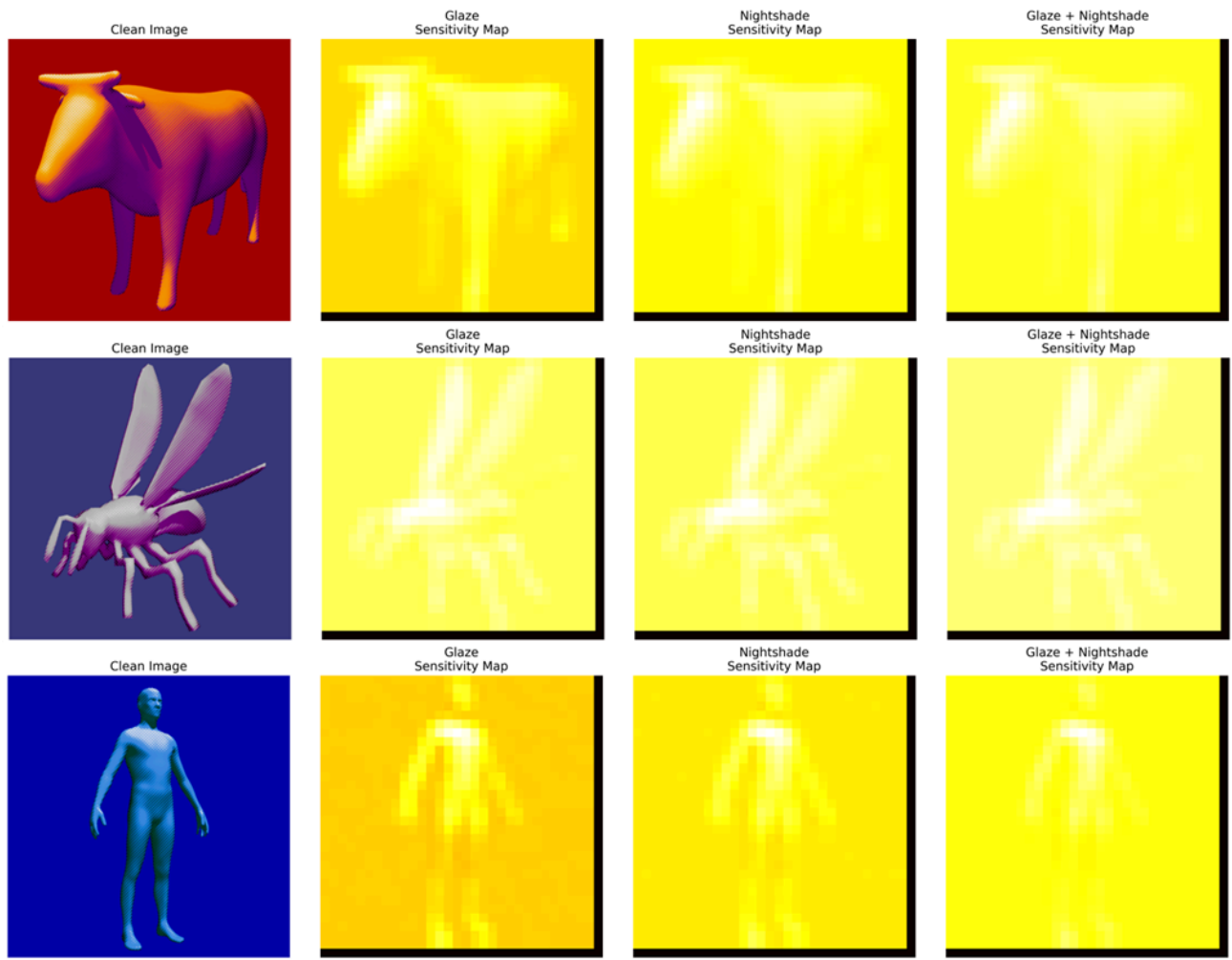

*Figure 11: Occlusion sensitivity maps for three 3D models (cow, wasp, and human models) under Glaze, Nightshade, and Nightshade-glaze perturbations. Across all models, the overall spatial patterns are highly similar, with differences appearing primarily as tonal rather than structural variations. Glaze tends to produce slightly more object-localized contrast, Nightshade yields smoother and more globally distributed responses, and the combined perturbation generally resembles Nightshade with reduced contrast.*

By contrast, Nightshade produces a broader and more globally distributed sensitivity profile. Its influence extends into background regions and reduces contrast between object boundaries and surrounding space. This smoother spatial footprint is consistent with Nightshade's objective of inducing semantic misalignment rather than localized texture-level disruption. Importantly, the combined Nightshade-glaze perturbation closely follows Nightshade's global spatial profile, reinforcing that under sequential application, the second method dominates the resulting spatial dependency. On smooth, low-texture objects such as the soft-shaded cube shown in Figure 12, occlusion sensitivity exposes important methodological limits. In these cases, Glaze, Nightshade, and their combination yield nearly identical spatial sensitivity patterns, all emphasizing illuminated faces and edge-transition regions driven primarily by the object's intrinsic shading gradients. Differences across perturbation methods manifest only as subtle tonal shifts rather than changes in spatial structure.

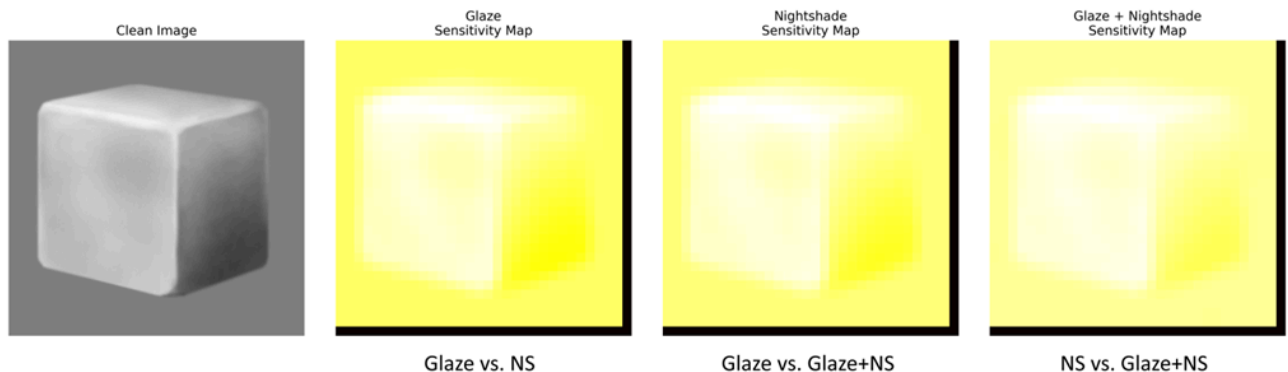

*Figure 12:* *Occlusion sensitivity maps for Glaze, Nightshade, and Nightshade-glaze on the soft-shaded cube. All three perturbations produce nearly identical spatial sensitivity patterns, with emphasis on the illuminated upper-left face and the cube's transition edges. Only minor tonal differences appear with slightly stronger contrast in Glaze, smoother gradients in Nightshade, and a uniformly bright response in the combined map.*

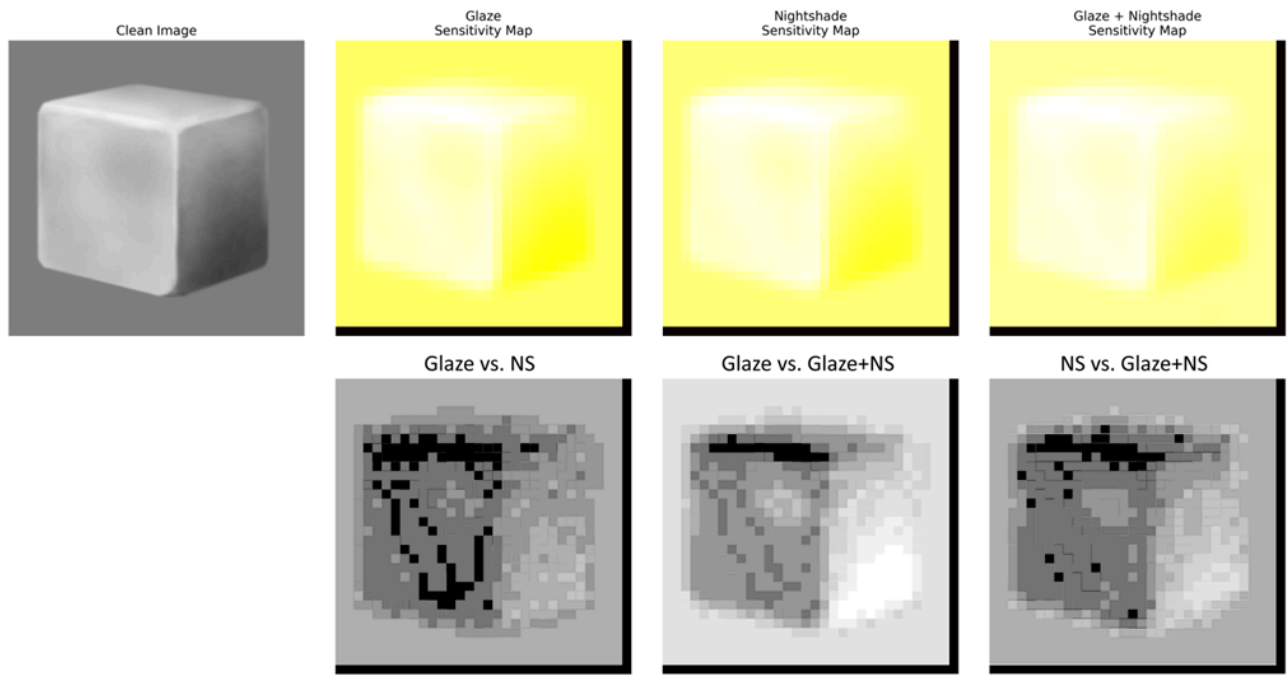

*Figure 13:* *Sensitivity maps and difference maps for Glaze, Nightshade, and Nightshade-glaze on a solid cube illustration from Figure 12. Raw sensitivity maps appear nearly identical across all three perturbations, revealing the limits of direct visualization. However, the difference maps expose where the methods diverge: Darker areas in grey difference maps show the least difference, while the brightest show the most difference between the compared perturbation methods. Glaze and Nightshade differ mainly on the cube's top and right faces. NS-Glaze remains largely Glaze-dominated with only localized Nightshade-driven shifts, and Nightshade versus NS-Glaze shows broad similarity with Glaze adding highlight-level structure.*

This behavior demonstrates that on uniform materials and low-frequency shading fields, occlusion sensitivity becomes dominated by the image's lighting distribution rather than by perturbation-specific structure. As a result, spatial heatmaps alone cannot reliably distinguish poisoning styles in such cases. To expose perturbation-specific deviations that are suppressed in raw occlusion maps, brightness-difference overlays between sensitivity maps were computed and contrast-compressed. As visualized in Figure 13, these difference visualizations reveal that although the baseline spatial patterns remain aligned, Glaze and Nightshade diverge most strongly on lighting-facing surfaces rather than shaded regions. In the cube example, the top and side faces exhibit the largest sensitivity differences, indicating that the two mechanisms modulate model response most distinctly along dominant illumination and gradient orientations. Further comparisons show that the Nightshade-glaze combination is largely Glaze-dominated with localized Nightshade-driven shifts, particularly in high-impact corner and edge regions. Conversely, Nightshade versus Nightshade-glaze exhibits broad similarity over illuminated surfaces, with Glaze introducing localized highlight-level modulation. These overlays reveal consistent compositional behavior: Nightshade establishes the global spatial diffusion pattern, while Glaze injects sharper, localized intensity modulation on top of it.

### 5.5 Frequency-Domain Signatures of Protection Mechanisms

While occlusion-based sensitivity analysis establishes that all protection methods remain spatially anchored to the clean image, its coarse spatial resolution and structure-dominated responses limit its ability to disentangle method-specific differences. To expose structural distinctions that are partially suppressed in the spatial domain, we turn to frequency-domain analysis. By examining Fourier log-magnitude spectra, signed spectral difference maps, and radially averaged frequency profiles, we obtain a model-agnostic characterization of how Glaze, Nightshade, and their sequential combination redistribute energy across spatial frequencies.

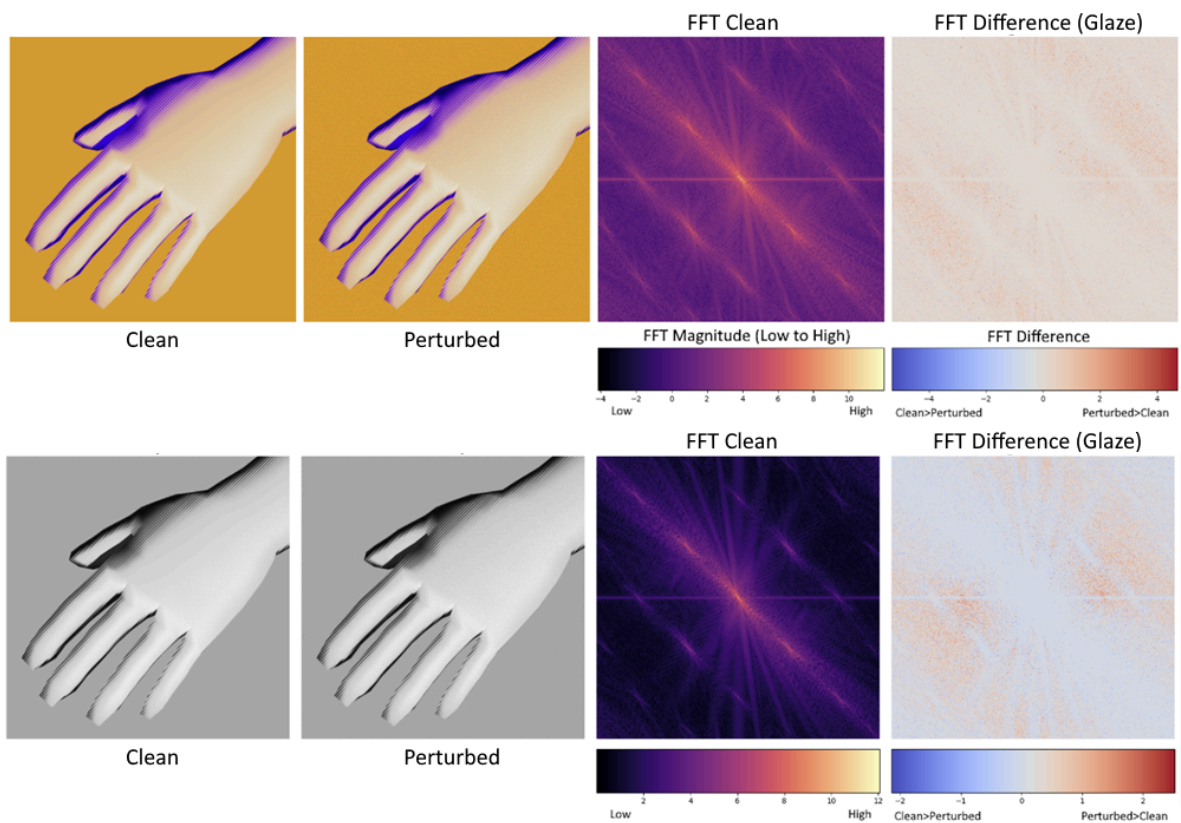

*Figure 14:* *FFT magnitude and difference maps for clean vs. Glaze-perturbed 3D hand. The RGB version shows diffuse, speckled differences across channels, while the grayscale FFT reveals a clear star-shaped structure with perturbation energy concentrated along the same oriented frequency bands present in the clean image. This illustrates how grayscale conversion suppresses color-channel noise and exposes the organized, directional nature of Glaze's frequency-space modifications.*

Across all evaluated images, grayscale Fourier analysis consistently yields more interpretable and structured perturbation signatures than RGB spectra. In the 3D hand example in Figure 14, the RGB Fourier difference map appears as a diffuse superposition of mixed positive and negative responses across channels, obscuring coherent structure. In contrast, the grayscale Fourier spectrum exposes a pronounced star-shaped pattern aligned with the dominant shading orientations of the object, and the corresponding signed difference map shows that perturbation energy concentrates precisely along these same directional frequency bands. This demonstrates that Glaze perturbations are not spectrally random but instead align with and amplify the intrinsic geometric frequency structure of the artwork. This behavior generalizes across object categories.

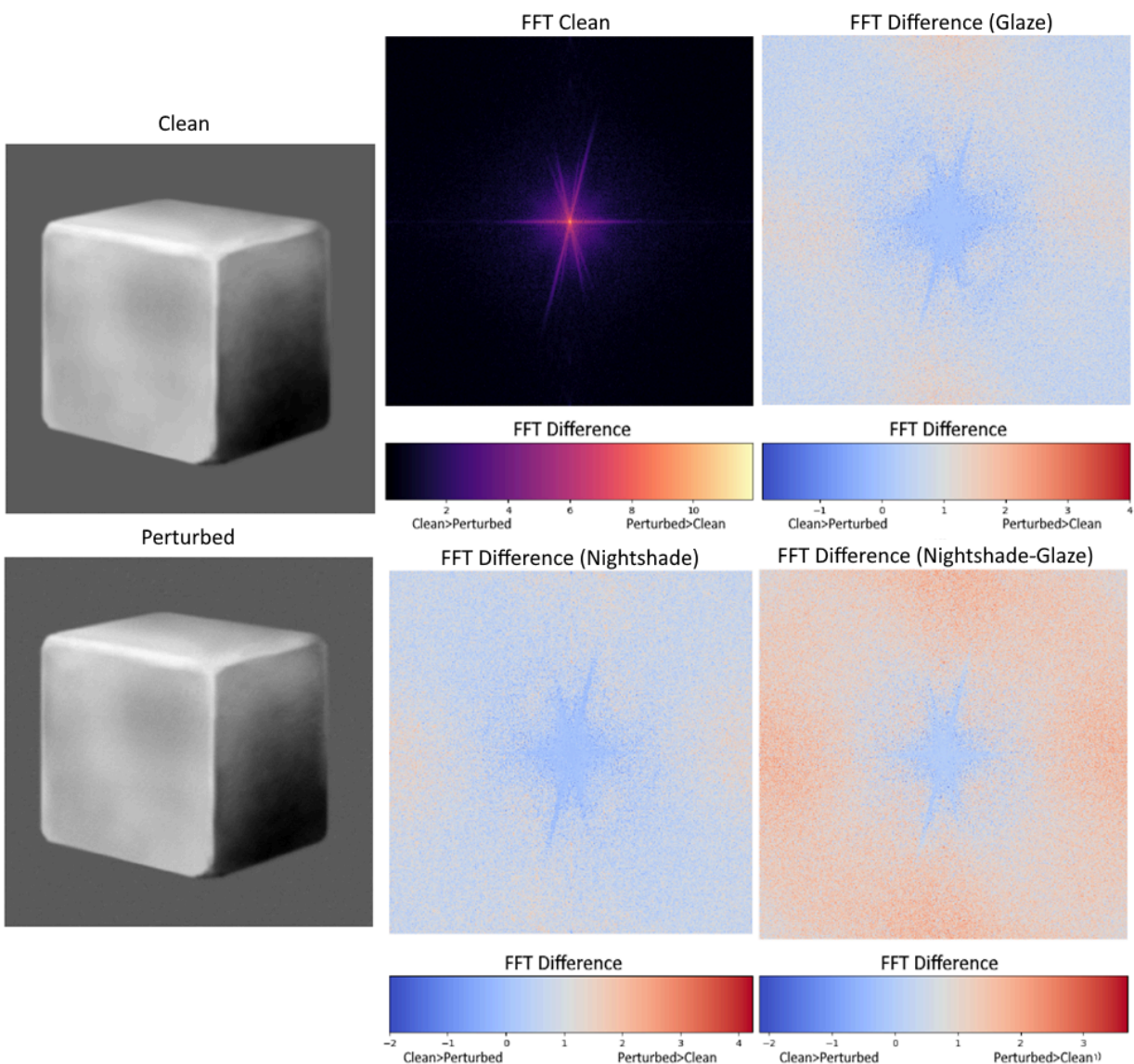

*Figure 15*: *The clean cube shows a strong glow-like FFT structure. Glaze's difference map contains more blue surrounding the center (indicating suppressed low-frequency energy) with only light red speckling at the periphery, reflecting modest, geometry-aligned high-frequency additions. Nightshade produces a nearly identical pattern. In contrast, the combined Nightshade-glaze map shows broader red regions, meaning the joint perturbation adds more overall spectral energy while still following the cube's inherent axial and diagonal structure (indicated by the blue center) rather than introducing random noise.*

In the soft-shaded cube example (Figure 15), the clean Fourier spectrum exhibits a strong central low-frequency concentration with axial and diagonal spokes corresponding to shading gradients. Both Glaze and Nightshade produce signed difference maps dominated by low-frequency suppression near the DC component (near the DC component corresponding to zero spatial frequency and global image mean; see Algorithm 3A) and modest, geometry-aligned high-frequency additions at the spectral periphery. In contrast, the combined Nightshade-glaze perturbation yields a broader red-dominant spectral field, indicating a larger overall increase in spectral energy while still respecting the cube's inherent axial and diagonal frequency structure. Importantly, none of the perturbations introduces spectrally uncorrelated noise; all energy redistributions remain geometrically organized.

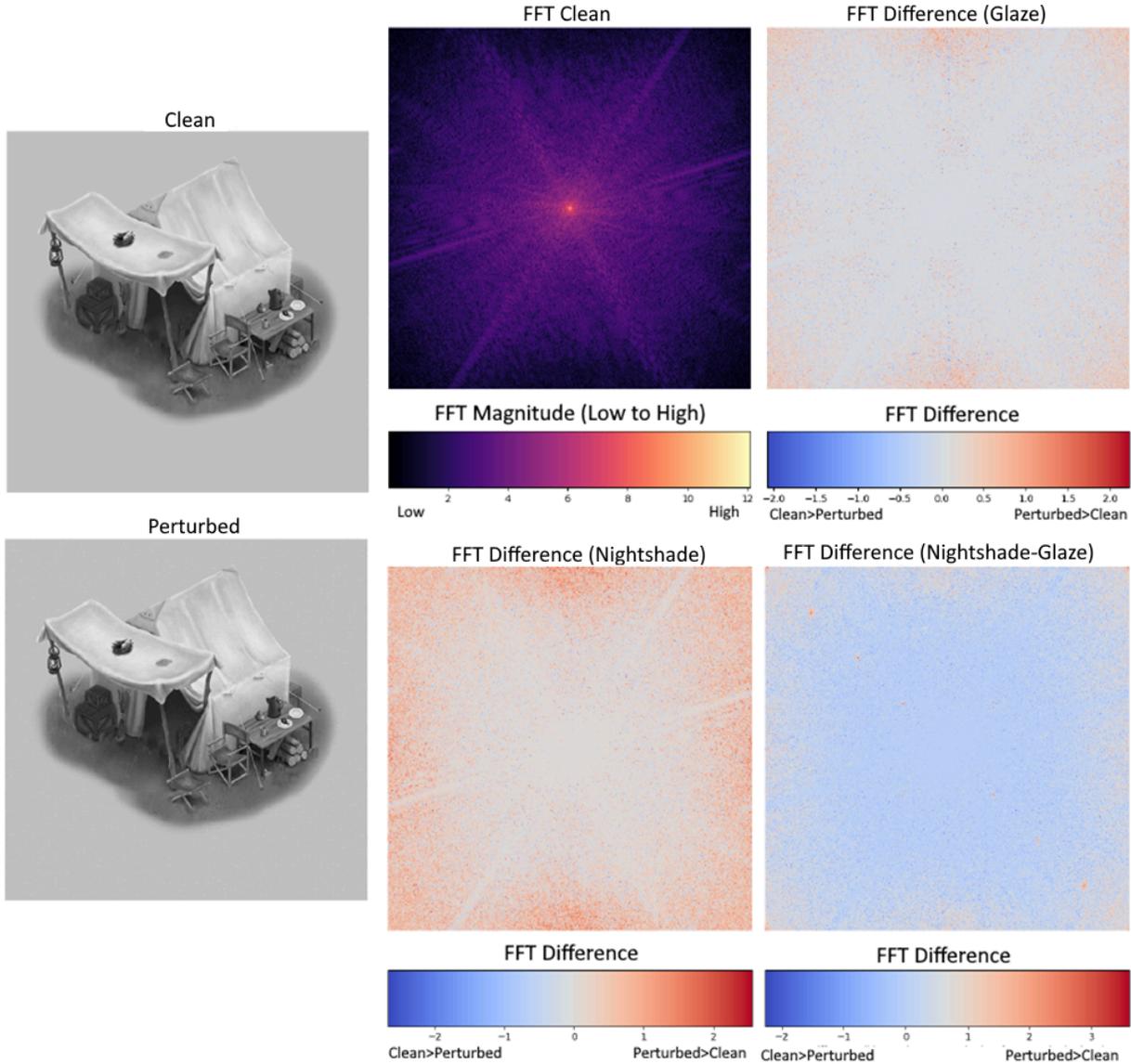

*Figure 16*: *The clean FFT for the tent scene exhibits a wider star pattern characteristic of its isometric shading and detailed edges with added detail. Once more, Glaze's difference map shows mostly faint blue with sparse red speckling, indicating slight suppression of low-frequency energy and mild, structure-aligned high-frequency additions. Nightshade produces a similar blue-dominant pattern. In contrast, the combined Nightshade-glaze perturbation yields a more uniformly red field, reflecting a larger overall increase in spectral energy while still preserving the scene's inherent directional structure rather than introducing random noise.*

A similar trend appears in the tent scene shown in Figure 16, where the clean FFT displays a wider star pattern characteristic of its isometric shading and detailed edge structure. Glaze and Nightshade once again exhibit predominantly blue-dominant difference maps with sparse red additions, reflecting slight low-frequency suppression and mild, structure-aligned high-frequency amplification. By contrast, the combined perturbation produces a more uniformly red spectral field, demonstrating that sequential application increases total spectral energy while preserving the scene's directional frequency organization rather than destroying it.

To complement these two-dimensional visualizations, radially averaged frequency profiles provide a direction-agnostic summary of global spectral redistribution (Figure 17). Across all examples, clean images exhibit the expected 1/f decay, with strong low-frequency energy tapering smoothly toward fine detail. Glaze shifts this curve upward in a globally coherent manner. At very low frequencies, the Glaze spectrum rises sharply above the clean baseline, indicating the introduction of subtle, broad, low-frequency fields rather than pixel-level noise. Across the mid-frequency band, the Glaze and clean curves remain approximately parallel, with Glaze consistently elevated, reflecting uniform addition of fine-scale texture energy.

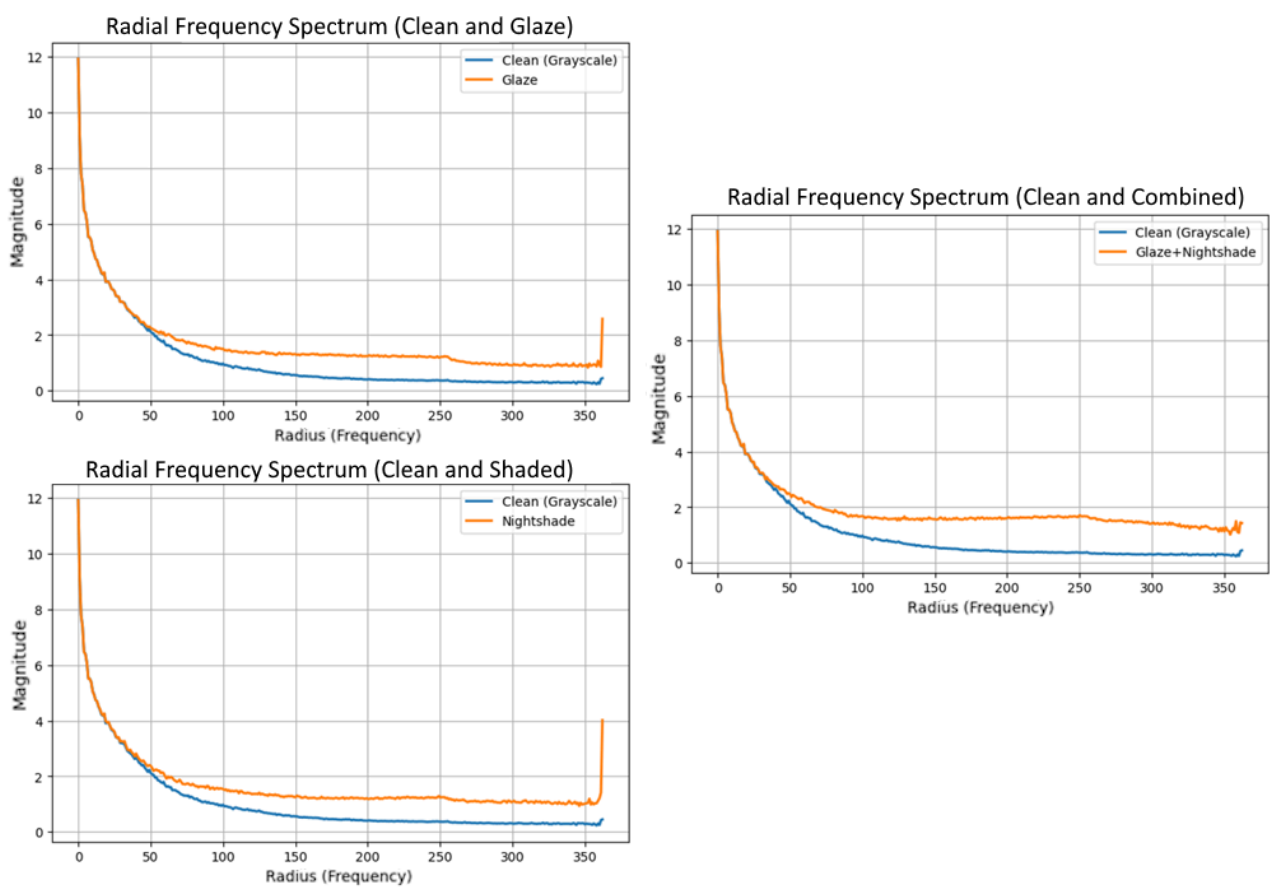

*Figure 17:* *Radial frequency spectra for Clean, Glaze, Nightshade, and Nightshade-glaze from the Solid cube illustration shown in Figure 15. All perturbations elevate the spectrum relative to the clean baseline, but with distinct signatures: Top Left: Glaze produces smooth global elevation. Bottom Left: Nightshade introduces stronger low-frequency boosts and a sharper high-frequency tail. Right: The combined Nightshade-glaze perturbation yields amplified low–mid frequencies with a more controlled high-frequency spike. These curves illustrate how each method, and especially their sequential combination, reshapes the spatial frequency distribution of the image.*

Nightshade exhibits a related but distinct spectral signature. While it also elevates the spectrum globally, it produces a stronger low-frequency boost and a sharper high-frequency tail, consistent with its gradient-driven perturbation mechanism that injects localized irregularities at fine spatial scales. When Glaze is applied after Nightshade, the resulting spectrum shows amplified low-to-mid frequency energy together with a noticeably smoother, reduced high-frequency spike. This hybrid signature indicates that Glaze partially regularizes the high-frequency irregularities introduced by Nightshade while reinforcing the broad structural energy injected earlier in the spectrum.

These frequency-domain results establish that modern protection methods do not operate as spectrally unstructured noise processes. Instead, all perturbations systematically redistribute energy along the dominant frequency orientations already present in the image. Glaze produces smooth, globally coherent spectral elevation; Nightshade contributes stronger low-frequency shifts and sharper high-frequency components; and their sequential combination yields a structured hybrid that inherits properties of both. This explains why perturbations remain visually subtle yet consistently detectable: they "ride along" the image's natural frequency axes rather than masking themselves as independent noise fields. This finding directly supports the low-entropy reconstruction assumption underlying LightShed's detection mechanism. By occupying predictable, structured frequency bands tied to the artwork's own spectral content, the perturbations remain compressible and recoverable by an autoencoder trained to isolate such low-entropy signals. The Fourier and radial analyses, therefore, provide a purely black-box confirmation that current protection mechanisms introduce systematic, frequency-aligned microstructure rather than spectrally diffuse randomness.

**5.6 Cross-Modal Synthesis of Representational, Spatial, and Spectral Evidence**
The analyses presented in Sections 5.1 through 5.5 characterize image protection mechanisms from complementary representational, activation, detectability, spatial, and frequency-domain perspectives. When viewed jointly, these results reveal a coherent and unified mechanism by which modern protection methods operate: perturbations are not random or diffuse artifacts, but instead remain tightly structured, low-entropy, and geometrically aligned with the underlying image content across all observable domains.

At the representational level, feature-space organization demonstrates that protection methods do not destroy semantic embedding structure, but instead introduce method-specific substructure within content-aligned clusters. This establishes that perturbations operate as constrained deformations within the image's existing feature manifold, rather than inducing global representational drift. The activation-level analysis then shows how these constrained perturbations are selectively amplified by stable subsets of internal feature channels, propagating perturbation structure across network depth while remaining coupled to original semantic content. The detectability results further refine this interpretation. Tool-based experiments demonstrate that semantic poisoning introduces strongly reconstructable and detectable perturbation signals, while stylistic camouflage partially suppresses detectability but does not eliminate the underlying structured signal. The controlled synthetic experiments reveal the foundational signal-level constraints: entropy magnitude, noise distribution, and spatial dispersion act as separable yet interacting controls over detectability, independent of semantic intent. Together, these findings establish that detectability is governed not merely by perturbation presence, but by whether the perturbation occupies compressible, structured signal subspaces that the purification model has learned to isolate.

Spatial occlusion analysis confirms that perturbation structure remains anchored to the geometry and shading of the clean image, rather than being spatially independent. Protection signals concentrate preferentially along edges, high-curvature regions, and illumination-driven gradients, with method-dependent differences emerging primarily as local contrast modulation rather than wholesale spatial relocation. This spatial anchoring explains why perturbations remain visually subtle while still producing consistent internal and spectral signatures. Frequency-domain analysis provides the final unifying constraint. All protection mechanisms redistribute spectral energy along the dominant frequency orientations already present in the image, rather than introducing spectrally unstructured noise. Glaze produces smooth, globally coherent spectral elevation; Nightshade contributes stronger low-frequency shifts and sharper high-frequency components; and their sequential combination yields a structured hybrid that inherits both behaviors. These frequency-aligned microstructures explain why perturbations remain both low-entropy and reconstructable, directly supporting the low-entropy assumption underlying LightShed's reconstruction-based detection strategy.

The combined white-box and black-box evidence establishes that contemporary image protection mechanisms operate through a shared structural principle: they embed perturbations that remain compressible, spatially anchored, and spectrally aligned with the artwork's own signal geometry. This unified structure explains their simultaneous visual subtlety, downstream model effectiveness, and detectability under reconstruction-based defenses. Rather than functioning as adversarial noise in the classical sense, these perturbations behave as structured, content-coupled micro-modulations that exploit both the geometry of natural images and the inductive biases of deep visual encoders.

## 6. Conclusion

This work presented a unified explainable AI framework for characterizing and interpreting adversarial image protection mechanisms through complementary feature-space, activation-level, detectability, spatial, and frequency-domain analyses. By integrating white-box and black-box perspectives, we provide a coherent mechanistic explanation of how contemporary protection methods operate, why they remain visually subtle, and under what conditions they remain detectable. With respect to **RQ1**, our embedding-space and activation analyses demonstrate that Glaze, Nightshade, and their sequential combination preserve content-driven feature organization while introducing structured, method-specific substructure. Protection mechanisms do not induce global representational drift; instead, they operate as constrained perturbations that remain tightly coupled to the image's semantic embedding, preserving base-image cluster structure with protection-specific subclustering, as shown in Section 5.1 and the associated t-SNE visualization. This establishes that modern image protection functions through structured feature-level deformation rather than semantic dislocation, which explains why detection models can still reliably identify object-level concepts in Nightshade-protected images even when generative models produce severely degraded outputs, as the perturbation aims to disrupt generative training dynamics rather than erase semantic structure at inference time. Addressing **RQ2**, our spatial occlusion and frequency-domain analyses reveal that perturbations are neither spatially independent nor spectrally diffuse. Instead, perturbation energy remains anchored to edges, high-curvature geometry, and illumination-aligned gradients, and is redistributed along the image's dominant frequency orientations rather than injected as unstructured noise. These findings explain how protection signals achieve high perceptual subtlety while maintaining consistent internal signatures that enable reconstruction-based detection. For **RQ3**, our tool-based and synthetic detectability experiments show that evasion is fundamentally constrained by signal structure rather than by semantic intent alone. Semantic poisoning remains strongly detectable due to its low-entropy, compressible structure, while stylistic camouflage partially suppresses detection but does not eliminate structured signal content. Controlled synthetic perturbations further demonstrate that entropy magnitude, spatial deployment, and noise distribution act as separable but interacting controls over detectability, revealing fundamental limits on evasion under reconstruction-based defenses. The cross-modal evidence establishes that modern image protection mechanisms operate through a shared structural principle: they embed low-entropy, spatially anchored, and frequency-aligned micro-modulations that exploit both the geometry of natural images and the inductive biases of deep visual encoders. This explains the simultaneous visual imperceptibility, downstream model interference, and detectability under explainable purification models observed across all experiments. Rather than behaving as classical adversarial noise, these perturbations function as content-coupled structural signals that persist across representational, spatial, and spectral domains.

*Limitations and Future Work* – This study focuses on a single reconstruction-based detection architecture and a fixed set of protection tools and synthetic perturbation families. While the cross-modal consistency of our findings suggests generality, extending this analysis to diverse purification architectures, diffusion-based detectors, and non-reconstruction-based defenses remains an important direction for future work. Additionally, while our frequency and spatial analyses expose the structural constraints of current protection methods, designing provably unpredictable, spectrally diffuse, and entropy-adaptive protection schemes represents a promising path toward stronger evasion under explainable defenses.

*Broader Implications* – By demonstrating that contemporary protection mechanisms remain structurally interpretable across multiple explanatory domains, this work reframes adversarial image protection not as an opaque cat-and-mouse game, but as a measurable signal-design problem governed by geometric and spectral constraints. These results provide a foundation for the development of next-generation protection strategies and detection models grounded in explainable, signal-level guarantees, with direct implications for generative model security, dataset ownership, and adversarial robustness in vision systems.

## Author Contributions